\documentclass[conference]{IEEEtran}
\usepackage{times}

\usepackage[numbers]{natbib}
\usepackage{multicol}
\usepackage[bookmarks=true]{hyperref}
\usepackage{xcolor}
\usepackage{amsmath}
\usepackage{amssymb}
\usepackage{algorithm}
\usepackage{algorithmicx}
\usepackage{graphicx}
\usepackage{wrapfig}
\usepackage{array}
\usepackage{xcolor}
\usepackage{booktabs}
\definecolor{CustomSkyBlue}{rgb}{0, 0.69, 0.94}
\definecolor{CustomPink}{rgb}{0.937, 0.580, 0.6196}
\usepackage{caption}
\usepackage{makecell}  
\usepackage{multirow}
\usepackage{adjustbox}
\usepackage{algorithm}          
\usepackage{algpseudocode}      
\usepackage{amsmath}            
\usepackage{amssymb}            
\usepackage{amsfonts}           
\usepackage{listings}
\usepackage{xcolor}
\usepackage{minted}
\definecolor{mintbg}{RGB}{248,248,248}
\definecolor{mintframe}{RGB}{210,210,210}
\let\oldtwocolumn\twocolumn
\renewcommand\twocolumn[1][]{
    \oldtwocolumn[{#1}{
	\begin{center}
    \vspace{-0.25in}
           \includegraphics[width=0.846\textwidth]{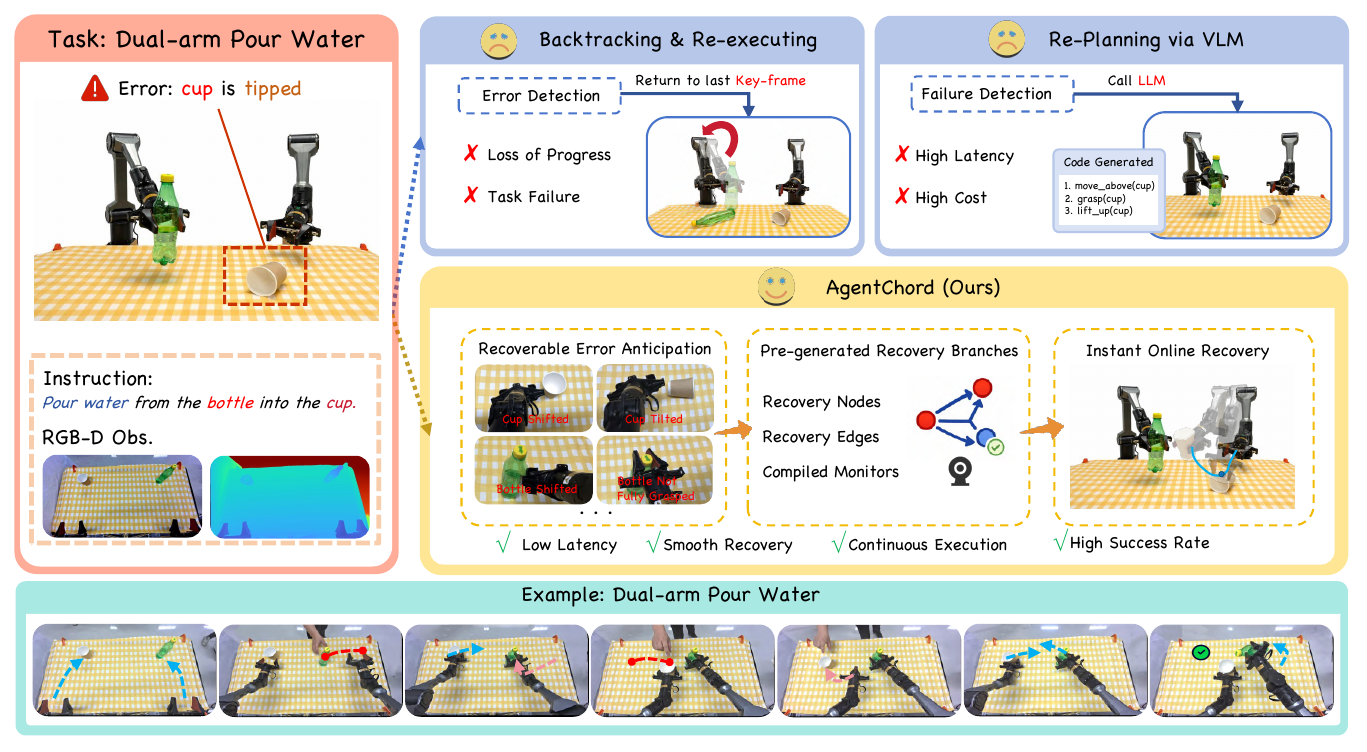}
           \vspace{-0.1in}
	\captionof{figure}{We introduce \textbf{AgentChord}, an agentic system that integrates anticipatory failure recovery capability, enabling both efficient and effective recovery actions. In the example, \textcolor{CustomSkyBlue}{blue} lines represent nominal actions, \textcolor{red}{red} lines indicate intrinsic failures or external disturbances, and \textcolor{CustomPink}{pink} lines depict recovery actions proactively anticipated by AgentChord.
    }\label{fig:teaser}
    \vspace{0.1in}
	\end{center}
    }]
}

\begin{document}

\title{From Reaction to Anticipation: Proactive Failure Recovery through Agentic Task Graph for \\Robotic Manipulation}

\author{
\authorblockN{
Sheng Xu$^{1}$,
Ruixing Jin$^{1}$,
Huayi Zhou$^{1}$,
Bo Yue$^{1}$,
Guanren Qiao$^{1}$,\\
Yueci Deng$^{1}$,
Yunxin Tai$^{2}$,
Kui Jia$^{1,2}$,
Guiliang Liu$^{1,3*}$
}
\authorblockA{
$^{1}$School of Data Science, The Chinese University of Hong Kong, Shenzhen\\
$^{2}$DexForce Technology,
$^{3}$Shenzhen Loop Area Institute.\\
$^{*}$Corresponding author: Guiliang Liu, \texttt{liuguiliang@cuhk.edu.cn}.
}
}

\maketitle
\begin{abstract}
Although robotic manipulation has made significant progress, reliable execution remains challenging because task failures are inevitable in dynamic and unstructured environments. To handle such failures, existing frameworks typically follow a stepwise detect-reason-recover pipeline, which often incurs high latency and limited robustness due to delayed reasoning and reactive planning. 
Inspired by the human capability to anticipate and proactively plan for potential failures, we introduce AgentChord, an agentic system that models a manipulation task as a directed task graph.
Before execution, this graph is enriched with anticipatory recovery branches that specify context-aware corrective behaviors, enabling immediate and targeted responses when failures occur. Specifically, AgentChord operates through a choreography of specialized agents: a composer that structures the nominal task graph, an arranger that augments the graph with anticipatory recovery branches, and a conductor that compiles and coordinates executable transitions using low-latency monitors to detect deviations and trigger pre-compiled recoveries without re-planning. Empirical studies on diverse long-horizon bimanual manipulation tasks demonstrate that AgentChord substantially improves success rates and execution efficiency, advancing the reliability and autonomy of real-world robotic systems. The project page is available at: \url{https://shengxu.net/AgentChord/}.
\end{abstract}

\IEEEpeerreviewmaketitle

\section{Introduction}
Robotic manipulation has been a central focus in the robotics community for many years. Over time, it has evolved from basic industrial tasks to more sophisticated operations in dynamic, unstructured environments~\cite{billard2019trends, li2025developments, mason2018toward, zheng2025survey, zhu2022challenges}. With advancements in artificial intelligence, robots are now capable of performing complex and adaptive tasks that require dexterity, flexibility, and decision-making, significantly broadening their potential applications and impact~\cite{fang2019survey, han2023survey, zhang2025generative}. However, despite these advancements, achieving reliable manipulation remains a significant challenge~\cite{bai2025towards, zhou2025yoto}. One of the major obstacles is the inevitability of failures during interactions with dynamic and unpredictable environments. As such, the ability to automatically detect and recover from failures is essential for the reliability of robotic systems in real-world settings~\cite{dai2025racer, duan2024aha, liu2023reflect, xia2025phoenix, yang2025fpc}. 

To \textit{identify potential failures}, recent studies have incorporated Multimodal Large Language Models (MLLMs) \cite{singh2025openai, awais2025foundation, carion2025sam, grattafiori2024llama, google2025gemini3} into robotic systems, framing failure detection as a specific case of the visual question answering (VQA) problem \cite{duan2024aha, liu2023reflect, yu2024multireact}.
Building on this, subsequent work proposes frameworks for \textit{failure recovery} by either synthesizing corrective actions from predefined skill libraries~\cite{guo2024doremi, huang2022inner} or generating language-based instructions to guide language-conditioned models~\cite{dai2025racer, lu2025robofac, xia2025phoenix, yang2025fpc}. However, directly deploying MLLMs introduces two key challenges~\cite{zhou2025code}: (i) slow inference caused by local computation and API-call latency, and (ii) imprecise grounding, as language-based instructions are coarse-grained without capturing details such as object positions, spatial relationships, exact angles, or distances. 

Striving for timely and accurate guidance, recent studies~\cite{huang2024rekep, zhou2025code} have leveraged MLLMs to extract key constraints, compile them into executable monitoring code, and verify numerical conditions during execution. Although this approach minimizes the need for frequent VQA calls and enables real-time monitoring, the generated monitoring code is often overly simplistic and limited to a small set of hand-crafted error types. More critically, the underlying failure recovery remains ineffective: existing methods either i) re-execute the entire pipeline after a failure~\cite{zhou2025code}, resulting in high latency and computational cost, or ii) utilize backtracking and re-execution strategies~\cite{huang2024rekep}, which tend to be unreliable in complex or perturbed bimanual manipulation scenarios.

Based on the previous analysis, we argue that the step-by-step ``detect-reason-recover'' pipeline remains ineffective for handling robotic failures in real time. In contrast, humans, as highly intelligent embodied agents, often {\it anticipate potential failures and formulate contingency plans in advance} during the process of manipulation. Once a failure occurs, parts of these plans are rapidly and often unconsciously activated, enabling swift recovery before the failure results in more severe consequences (e.g., stabilizing a tipping bottle of water before it spills across the table).

To achieve human-level dexterity in failure recovery, we propose proactively generating recovery actions alongside failure anticipation, 
thereby enabling immediate and targeted correction when failures occur. 
To this end, we introduce \textbf{AgentChord}, a choreographed agentic system that tightly integrates failure anticipation, detection, and recovery within a unified framework. Under this design, AgentChord models a manipulation task as a directed graph, where nodes encode semantic sub-goals and edges represent motion transitions. Crucially, the graph is {\it augmented before execution} with recovery branches that define corrective actions for specific failures, allowing quick transitions to a recovery path that maintains task progress. These branches are forward-moving, ensuring continuous progress toward the goal without regression.

From a system perspective, AgentChord operates through a choreography of specialized agents whose roles mirror those in a musical ensemble. The \textbf{task structuring agent} acts as a \emph{composer}, interpreting high-level language instructions and visual observations to compose a structured task graph with explicit sub-goals. The \textbf{recovery orchestration agent} plays the role of an \emph{arranger}, augmenting this nominal graph with anticipatory recovery branches by reasoning about likely failure modes and inserting recovery nodes, recovery transitions, and merge points. The \textbf{execution compilation agent} functions as a \emph{conductor}, transforming both nominal and recovery transitions into executable, interruptible programs via hierarchical constrained solvers that adapt the abstract graph to the robot’s embodiment, while coordinating online execution through compiled monitors. During execution, these monitors continuously evaluate multimodal constraint violations and trigger immediate transitions to pre-compiled recovery behaviors, enabling low-latency recovery without re-planning or regression to earlier stages.

Our contributions are summarized as follows: 1) We introduce \textbf{AgentChord}, an agentic framework that models manipulation tasks as recovery-augmented graphs, enabling unified task structuring, execution compilation, failure anticipation, and online recovery in an interpretable representation. 2) We propose a proactive, forward-moving recovery paradigm that embeds pre-determined recovery behaviors into the task graph, preventing regressive resets and re-planning, and develop a low-latency execution mechanism using compiled monitors and constrained solvers to enable immediate recovery transitions during execution. 3) We demonstrate that AgentChord achieves the highest average success rate and execution efficiency on diverse long-horizon bimanual manipulation tasks in both simulation and real-world environments, with its generated data supporting robust policy learning.

\section{Related Works}
\subsection{Agentic Systems for Robotic Manipulation}
With the rapid advancement of foundation models, researchers have increasingly applied them to robotic manipulation tasks~\cite{firoozi2025foundation, kawaharazuka2024real, li2024foundation}. Recent studies show that integrating large pre-trained models into robotic systems substantially improves perception, decision-making, and control, with growing interest in agentic frameworks~\cite{huang2023voxposer, zhou2025yoto}. In such systems, perception agents powered by Vision-Language Models (VLMs), including CLIP~\cite{radford2021learning}, Grounding DINO~\cite{liu2024grounding}, and SAM3~\cite{carion2025sam}, enable open-vocabulary recognition and segmentation for flexible object manipulation in dynamic environments. Reasoning and planning agents based on Large Language Models (LLMs) or code-generation frameworks such as ProgPrompt~\cite{singh2022progprompt} and Code-as-Policy~\cite{liang2022code} further allow robots to generate task plans and actions from high-level language instructions. Beyond using a single foundation model, recent work increasingly focuses on coordinating multiple foundation models as distinct agents~\cite{chen2025robotwin, duan2024manipulate, elmallah2025score, gong2025anytask, huang2024rekep, liu2025simpact, raptis2025agentic, yang2025maniagent, zhou2025code}, enabling more structured perception, planning, and execution. Some approaches additionally introduce VLMs as validation agents to supervise execution and improve robustness~\cite{duan2024manipulate, huang2024rekep, yang2025maniagent, zhou2025code}. However, existing methods do not fully exploit foundation models within a cohesive agentic framework that supports efficient and proactive failure anticipation and recovery.

\subsection{Failure Detection and Recovery in Robotic Manipulation}
To enable autonomous failure detection and recovery in unstructured environments, recent work has leveraged foundation models, particularly MLLMs. Most approaches frame failure detection as a VQA problem, where a VLM determines whether a failure has occurred from visual observations~\cite{du2023vision, duan2024aha, liu2024self, liu2023reflect, wang2023describe, yu2024multireact, zheng2024evaluating}. Building on this capability, subsequent studies propose automatic recovery frameworks that either generate executable corrective actions~\cite{guo2024doremi, huang2022inner} or use language-based instructions to guide language-conditioned models~\cite{dai2025racer, lu2025robofac, xia2025phoenix}. Beyond reactive recovery, recent work explores proactive failure detection~\cite{yang2025fpc, zhou2025code}. However, repeated MLLM invocations are computationally expensive, slow, and often yield coarse-grained detection. To address this, ReKep and Code-as-Monitor~\cite{huang2024rekep, zhou2025code} use foundation models to extract constraint points and compile detection code, enabling more efficient and precise monitoring. Nevertheless, their detectors remain tightly coupled to specific constraint points, limiting flexibility and generalization, and recovery is typically handled through full pipeline re-invocation or simple backtracking. In contrast, we propose to proactively generate recovery actions alongside failure anticipation, enabling systematic and efficient recovery without re-planning.

\section{Method}
\label{sec:method}
\begin{figure*}[htbp]
    \centering
    \includegraphics[width=\linewidth]{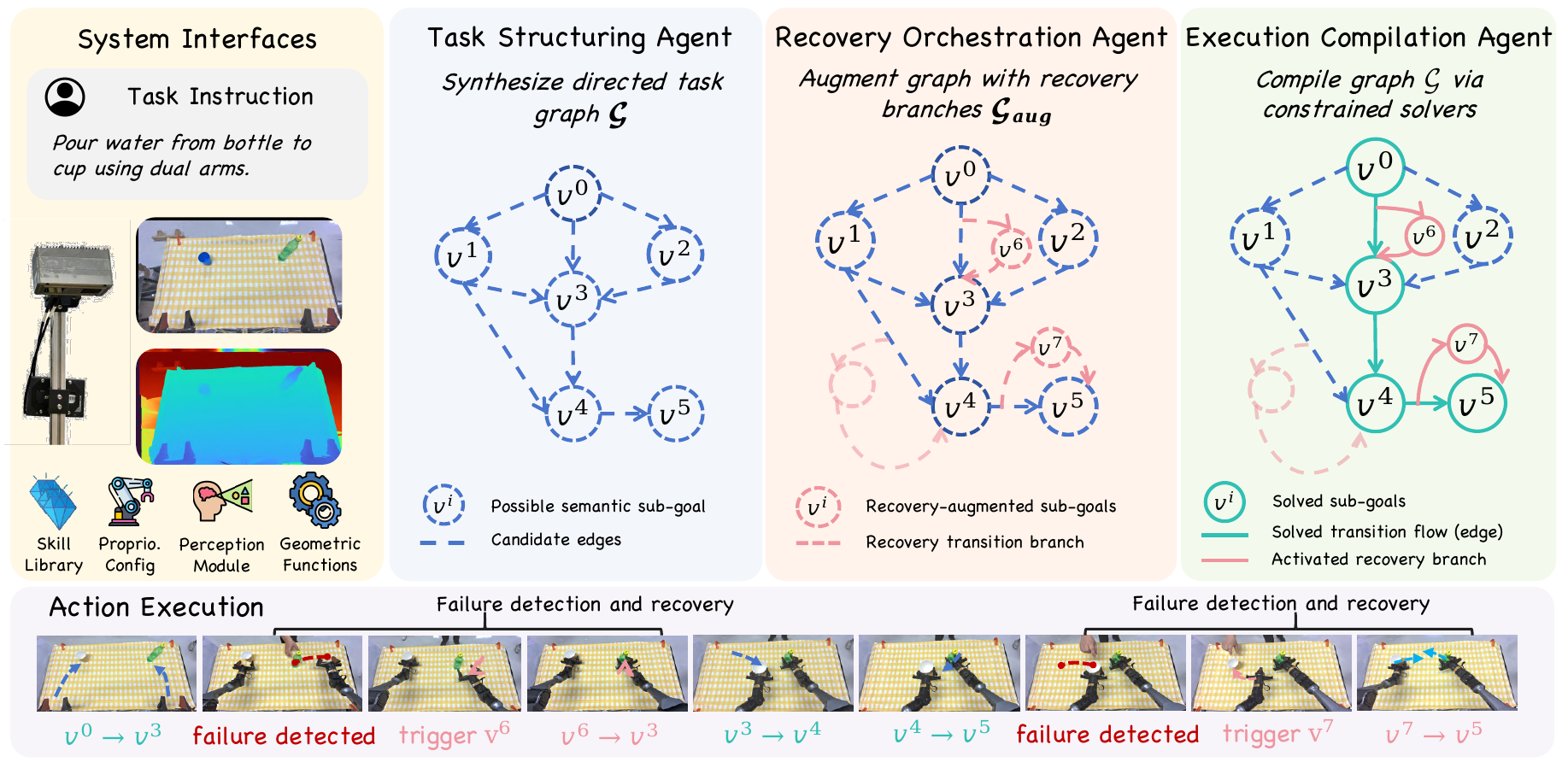}
    \vspace{-0.22in}
    \caption{The framework of \textbf{AgentChord} with an example. It involves three agents: the Task Structuring Agent constructs the nominal task graph with semantic sub-goals and constraints; the Recovery Orchestration Agent augments this graph with anticipatory recovery branches; and the Execution Compilation Agent compiles both nominal and recovery transitions into executable, interruptible programs with monitors. During execution, when a failure is detected, the system immediately switches to the corresponding pre-compiled recovery branch, enabling efficient recovery without re-invoking full task planning.}
    \vspace{-0.1in}
    \label{fig:framework}
\end{figure*}

In this section, we present the method of AgentChord. We first introduce the system interfaces and task-dependent feature representation (Sec.~\ref{sec:problem-formulation}). We then describe how AgentChord constructs a directed task graph (Sec.~\ref{sec:task-graph}), augments it with anticipatory recovery branches (Sec.~\ref{sec:failure-recovery}), and compiles the recovery-augmented graph into executable programs with monitors and triggers (Sec.~\ref{sec:solvers-code}). Finally, we summarize the agentic coordination and online execution workflow (Sec.~\ref{sec:workflow}). Fig.~\ref{fig:framework} illustrates the overall framework.

\subsection{Problem Formulation and System Interfaces}
\label{sec:problem-formulation}

\textbf{Functional tools.}
AgentChord is implemented as an agentic system with (i) perception tools $\mathcal{F}_{\text{perc}}$ (e.g., SAM3~\cite{carion2025sam}, AnyGrasp~\cite{fang2023anygrasp}, and FoundationPose~\cite{wen2024foundationpose}), (ii) an action library $\mathcal{F}_{\text{action}}$ of atomic manipulation skills (e.g., grasp, move relative to an object frame), and (iii) model-call tools $\mathcal{F}_{\text{mllm}}$ for invoking multimodal foundation models such as GPT~\cite{singh2025openai} or Gemini~\cite{google2025gemini3} to reason about the scene and task.

\textbf{Perception modules.}
Let $o_t$ denote the RGB-D observations captured by binocular cameras at timestep $t$. Given an observation $o_t$ and a queried object description for object $k$, we apply open-vocabulary segmentation using SAM3~\cite{carion2025sam} to obtain an object mask $m_t^k$. Fusing this mask with the corresponding depth information produces an object-level point cloud $\mathcal{P}_t^{k} \in \mathbb{R}^{N_k \times 3}$. We also integrate a grasp pose generator, AnyGrasp~\cite{fang2023anygrasp}, which outputs a high-confidence 6-DoF grasp pose for the target object $k$, enabling direct execution of grasping actions, which can be used by downstream edge programs. We use $\Psi(\cdot)$ to denote the perception modules.

\textbf{Robot state and measurable signals.}
Let $\xi_t = (q_t, g_t)$ denote the robot's measurable state at timestep $t$, where $q_t$ is the joint configuration and $g_t$ is the gripper state (e.g., finger distance or binary open/close states). Let $e_t$ represent the end-effector pose command, with $e_t = (e_t^L, e_t^R) \in SE(3)^2$ for bimanual manipulation and $e_t \in SE(3)$ for single-arm tasks. The robot configuration consistent with $e_t$ is denoted as $q(e_t)$ (e.g., obtained via inverse kinematics). These signals are used for both low-level control and failure detection.

\textbf{Geometric feature extraction.}
For constraint evaluation and monitoring, we define a task-dependent feature vector:
\begin{equation}
z_t = \Phi(\hat{x}_t, \xi_t),
\label{eq:rel_feature}
\end{equation}
where $\hat{x}_t = \Psi(o_t)$ denotes the perception output and $\xi_t$ is the robot's measurable state. In practice, $\hat{x}_t$ may include object masks, object-level point clouds, grasp poses, and compact geometric attributes derived from RGB-D observations. Given an object mask $m_t^k$ derived from $\mathcal{F}_{\text{perc}}$, we can fuse it with depth to obtain an object point cloud $\mathcal{P}_t^k$, from which $\Phi(\cdot)$ extracts task-relevant geometric quantities, such as object centroids, boundary or extremal points, principal orientation axes estimated from PCA, and fractional locations along an object, e.g., the center or quarter position of a bottle. The feature vector also incorporates robot-side signals from $\xi_t=(q_t,g_t)$, including the joint configuration, gripper opening width, and binary open/closed state, which are useful for detecting interaction states such as grasp closure or object attachment. Unlike previous research~\cite{huang2024rekep, zhou2025code} that explicitly relies on extracted constraint points from VLMs as the feature form, AgentChord does not require a fixed intermediate representation. Instead, it can leverage various forms of $z_t$ derived from perception and robot signals as required by the detection procedure, greatly enhancing flexibility and generalizability. More details are provided in Appendix~\ref{sec:appendix_geo_feat}.

\subsection{Directed Graph for Task Execution}\label{sec:task-graph}
AgentChord initializes the task graph using a \textbf{task structuring agent}. Specifically, given a task instruction $\mathcal{I}$ and an initial observation $o_0$, the agent invokes $\mathcal{F}_{\text{mllm}}$ with task-specific prompts to synthesize the directed task graph $\mathcal{G}$, including the node set, graph topology, and constraint specifications:
\begin{equation}
\mathcal{G} = (V, E),
\label{eq:graph}
\end{equation}
where each node $v^{i} \in V$ denotes a semantic sub-goal, and each directed edge $\varepsilon^{i\rightarrow j}\in E$ denotes a motion transition intended to move from $v^{i}$ to $v^{j}$ under constraints. Note that we use superscripts $i$ to denote a specific keyframe with meaningful semantics, while subscripts $t$ are used to represent individual time steps. This representation supports branching and merging, which is essential for encoding multiple feasible strategies and recovery routes.

\textbf{Constraint semantics.}
Each node $v^{i}$ is associated with sub-goal constraints $\mathcal{C}_{\text{sub}}^{i}$ describing successful completion at $v^{i}$. Each edge $\varepsilon^{i\rightarrow j}$ is associated with path constraints $\mathcal{C}_{\text{path}}^{i\rightarrow j}$ that should remain valid during the transition from $v^i$ to $v^j$. We write both types of constraints as inequality functions:
\begin{equation}
c(z) \le 0,
\label{eq:constraint_form}
\end{equation}
where $c(\cdot)$ denotes a constraint function over task-dependent features. For sub-goal constraints, $z$ is evaluated at the target node feature, e.g., $z^i$ for $v^i$. For path constraints, $z$ is evaluated along the transition, e.g., $z_t$ during execution. This design makes both node completion and edge execution constraints convenient to evaluate online and naturally compatible with compiled monitors.

\subsection{Proactive Failure Anticipation and Recovery}\label{sec:failure-recovery}
AgentChord attaches recovery behaviors to the nominal task graph \emph{before} execution. It utilizes a \textbf{recovery orchestration agent} to invoke $\mathcal{F}_{\text{mllm}}$ on $\mathcal{G}$, analyze each node and its outgoing edges, and predict a set of likely failure modes:
\begin{equation}
\mathcal{F}^{i\rightarrow j} = \{ f^{i\rightarrow j}_1, f^{i\rightarrow j}_2, \dots \}.
\label{eq:failure_set}
\end{equation}
Each failure mode $f^{i\rightarrow j}_m$ is paired with an online-detectable trigger and a recovery intent (e.g., re-align, re-approach, or re-grasp). The perception modules provide visual cues such as object pose drift, relative distances or angles, and the robot provides gripper-state cues such as finger distance or grasp closure. Both are integrated in $z_t$ via Eq.~\eqref{eq:rel_feature} and are directly usable for online failure detection.

\textbf{Graph augmentation with recovery branches.}
For each predicted failure mode $f^{i\rightarrow j}_m$, AgentChord specifies a recovery node 
$v_m^{\text{rec}(i,j)}$ together with recovery edges that first route execution from the failure 
context to the recovery node and then merge it back into a feasible downstream node. 
Over all feasible nominal edges $\varepsilon^{i\rightarrow j}\in E$, the set of recovery nodes is defined as:
\begin{equation}
V_{\text{rec}}
=
\left\{
v_m^{\text{rec}(i,j)}
\;\middle|\; 
f_m^{i\rightarrow j}\in \mathcal{F}^{i\rightarrow j}
\right\}.
\label{eq:recovery_nodes}
\end{equation}
Let $K_m^{i\rightarrow j}\subseteq V$ denote the set of candidate downstream merge 
nodes for the recovery node $v_m^{\text{rec}(i,j)}$. The corresponding recovery 
edge set is defined as:
\begin{equation}
\begin{aligned}
E_{\text{rec}}
=
&\left\{
\varepsilon^{i\rightarrow \text{rec}(i,j,m)}
\;\middle|\;
f_m^{i\rightarrow j}\in \mathcal{F}^{i\rightarrow j}
\right\}
\\
&\cup
\left\{
\varepsilon^{\text{rec}(i,j,m)\rightarrow k}
\;\middle|\;
f_m^{i\rightarrow j}\in \mathcal{F}^{i\rightarrow j},\ 
v^k\in K_m^{i\rightarrow j}
\right\}.
\end{aligned}
\label{eq:recovery_edges}
\end{equation}
Here, $\varepsilon^{i\rightarrow \text{rec}(i,j,m)}$ denotes the transition from 
the failure context to the recovery node $v_m^{\text{rec}(i,j)}$, while 
$\varepsilon^{\text{rec}(i,j,m)\rightarrow k}$ denotes the transition from this 
recovery node to a feasible downstream merge node $v^k$. The augmented graph is then:
\begin{equation}
\mathcal{G}_{\text{aug}} = (V \cup V_{\text{rec}},\ E \cup E_{\text{rec}}).
\label{eq:graph_aug}
\end{equation}

As with nominal graph elements, each recovery node is assigned sub-goal constraints, and each recovery edge is assigned path constraints. Notably, when a recovery intent aligns with the semantics and constraints of an existing sub-goal, AgentChord can directly route the recovery transition to the corresponding nominal node. In such cases, the recovery branch re-enters the nominal task graph without introducing additional nodes. Explicit recovery nodes are only instantiated when the recovery behavior requires objectives that differ from those of existing sub-goals. More details about failure anticipation and recovery are provided in Appendix~\ref{sec:appendix_failure_recovery}.

\textbf{Forward-moving recovery principle.}
To prevent regressive behavior, AgentChord constrains recovery to be forward-moving in the task graph. Let $\mathrm{Dist}(u,v)$ denote the shortest-path cost required to traverse from sub-goal $u$ to sub-goal $v$ in the augmented graph $\mathcal{G}_{\text{aug}}$, with edge weights defined by physical execution costs such as execution steps. When a violation is detected during execution of edge $\varepsilon^{i\rightarrow j}$, we define the failure context as $v^{\text{fail}}:= v^{i}$ (the last completed sub-goal at the time of failure). For any recovery node $v_m^{\text{rec}(i,j)}$ that may be triggered from this context, AgentChord enforces:
\begin{equation}
\mathrm{Dist}\!\left(v_m^{\text{rec}(i,j)}, v^N\right)
\le
\mathrm{Dist}\!\left(v^{\text{fail}}, v^N\right),
\label{eq:forward_progress}
\end{equation}
where $v^N$ denotes the terminal goal node. Eq.~\eqref{eq:forward_progress} enforces forward progress by ensuring that recovery does not increase the remaining graph-theoretic distance to the goal. In practice, the recovery orchestration agent first proposes candidate merge targets, after which Eq.~\eqref{eq:forward_progress} filters out branches whose remaining cost-to-go exceeds that of the failure state.

\subsection{Graph Compilation via Hierarchical Constrained Solvers}\label{sec:solvers-code}
AgentChord compiles the recovery-augmented graph into executable behaviors with an \textbf{execution compilation agent} by separating \emph{sub-goal realization} (node-level) from \emph{transition realization} (edge-level). This two-level structure parallels the common goal-vs-path decomposition~\cite{caval2014keeping, dietterich2000hierarchical, huang2024rekep}, and in AgentChord it is applied uniformly to both nominal transitions and recovery transitions.

\textbf{Node synthesis (sub-goal realization).}
Inspired by~\cite{huang2024rekep}, for each node $v^{i}$, AgentChord computes a target end-effector pose $e^{i}$ that satisfies the sub-goal constraints $\mathcal{C}_{\text{sub}}^{i}$ under the perceptual estimate $\hat{x}^{i}$ by solving the following problem:
\begin{align}
e^{i}
&= \arg\min_{e} \ \lambda_{\text{sub}}^{i}(e;\hat{x}^i) \nonumber\\
&\text{s.t.}\quad
c\!\left(\Phi(\hat{x}^i, (q(e), g^i))\right) \le 0,\ 
\forall c \in \mathcal{C}_{\text{sub}}^{i} .
\label{eq:subgoal_solver}
\end{align}
Here, $\hat{x}^i$ denotes the perceptual estimate used when synthesizing node $v^i$, $g^i$ denotes the intended gripper state at $v^{i}$ (e.g., open for pre-grasp, closed for post-grasp), and $\lambda_{\text{sub}}^{i}$ represents the optimization objectives, such as collision avoidance, reachability margins, and bimanual coordination regularizers. Eq.~\eqref{eq:subgoal_solver} computes a concrete keyframe pose for each node.

\textbf{Edge synthesis (transition realization).}
For each directed edge $\varepsilon^{i\rightarrow j}$, AgentChord
executes a constraint-aware transition strategy that drives the
robot from $e^i$ toward $e^j$ while satisfying path constraints
$\mathcal{C}_{\text{path}}^{i\rightarrow j}$. These path constraints
encode the conditions required for feasible and safe motion
generation, such as collision avoidance, reachability, and task-relevant geometric relations that should be maintained
during the transition. To remain robust under disturbances, AgentChord uses receding-horizon refinement. At each control step $t$, we solve the following problem:
\begin{align}
\mathbf{e}_{t:t+H}^{*}
&= \arg\min_{\mathbf{e}_{t:t+H}} \ 
\lambda_{\text{path}}^{i\rightarrow j}(\mathbf{e}_{t:t+H}; \hat{x}_t)
\label{eq:path_solver} \\
\text{s.t.}\quad
&c\!\left(\Phi\!\left(\hat{x}_{t}, (q(e_{t+h}), g_{t+h})\right)\right) \le 0, \nonumber\\
&\forall c \in \mathcal{C}_{\text{path}}^{i\rightarrow j}, \quad
\forall h \in \{0,\dots,H\}. \nonumber
\end{align}
Here, $\mathbf{e}_{t:t+H}=\{e_t,\dots,e_{t+H}\}$ denotes the planned end-effector pose sequence, and $g_{t+h}$ denotes the corresponding planned gripper state at step $t+h$. Since future observations are not directly available, the solver evaluates path constraints using the latest perceptual estimate $\hat{x}_t$ within each horizon, which is updated after every replanning step. We re-solve Eq.~\eqref{eq:path_solver} every $M$ control steps and execute only the first $M$ commands $\{e_t^{*}, e_{t+1}^{*}, \dots, e_{t+M-1}^{*}\}$ before replanning, where $M \le H$. The objective $\lambda_{\text{path}}^{i\rightarrow j}$ includes progress-to-goal terms (toward $e^j$), smoothness, and safety margins. In practice, we solve these optimization problems using the solvers implemented in EmbodiChain~\cite{EmbodiChain}.

\textbf{Compiled monitors and triggers.}
AgentChord separates the constraints used for action synthesis from those used
for failure-triggered recovery. The path constraints
$\mathcal{C}_{\text{path}}^{i\rightarrow j}$ guide the edge solver in generating
feasible transition actions, whereas recovery is triggered by the failure
functions associated with the active edge. Specifically, for an edge $\varepsilon^{i\rightarrow j}$ leaving node $v^i$, each anticipated failure mode
$f^{i\rightarrow j}_m\in\mathcal{F}^{i\rightarrow j}$ is represented by an executable scalar function over the feature vector $z_t$:
\begin{equation}
f^{i\rightarrow j}_m(z_t) \leq 0 .
\end{equation}
By convention, $f^{i\rightarrow j}_m(z_t)\leq 0$ indicates normal execution with respect to this failure mode, whereas $f^{i\rightarrow j}_m(z_t)>0$ indicates that the corresponding failure condition, such as object displacement, object tilt, loss of grasp, or relational misalignment, has been detected.

During execution, AgentChord continuously evaluates these failure-monitoring functions.
To avoid spurious activation caused by transient perception noise or short-lived
contact fluctuations, the monitor for failure mode $f^{i\rightarrow j}_m$ is activated only
when the function remains violated beyond a robustness margin for a short
persistence window:
\begin{equation}
\mathcal{M}^{i\rightarrow j}_m(t)
=
\mathbb{I}\!\left[
f^{i\rightarrow j}_m(z_{t'}) > \epsilon,\ 
\forall t' \in \{t\!\!-\!\!K\!\!+\!\!1,\dots,t\}
\right],
\label{eq:monitor}
\end{equation}
where $\epsilon\geq 0$ is a robustness margin and $K$ is the persistence window.
When $\mathcal{M}^{i\rightarrow j}_m(t)=1$, AgentChord switches execution according to the recovery mapping for the detected failure:
\begin{equation}
\rho\!\left(\varepsilon^{i\rightarrow j}, f^{i\rightarrow j}_m\right)
=
\varepsilon^{i\rightarrow \text{rec}(i,j,m)},
\label{eq:recovery_mapping}
\end{equation}
and immediately executes the corresponding pre-compiled recovery edge in
$\mathcal{G}_{\text{aug}}$. This design allows AgentChord to activate
failure-specific recovery behaviors without re-invoking full task planning.

\subsection{Agentic Coordination and Execution Workflow}
\label{sec:workflow}
AgentChord coordinates three agents with complementary roles. The \textit{task structuring agent} constructs the base task graph $\mathcal{G}$ together with its semantic sub-goals and constraints. The \textit{recovery orchestration agent} augments $\mathcal{G}$ into a recovery-augmented graph $\mathcal{G}_{\text{aug}}$ by adding anticipatory recovery nodes, recovery edges, and candidate merge targets. The \textit{execution compilation agent} then compiles both nominal and recovery transitions into executable, interruptible edge programs with the corresponding monitors. During execution, AgentChord traverses $\mathcal{G}_{\text{aug}}$, running the active edge program while evaluating the monitor. If a monitor triggers, execution switches to the recovery edge; otherwise, it continues to the next node when the target constraints are met. The implementation details are provided in Appendix~\ref{sec:imp-details}, and Algorithm~\ref{alg:agentchord} summarizes the complete AgentChord pipeline.

\section{Empirical Evaluation}
In this section, we conduct empirical evaluations in both simulation and real-world environments to answer the following research questions: 1) Can AgentChord autonomously solve diverse robotic manipulation tasks, demonstrating higher task success rates and reduced execution times by leveraging efficient failure detection and recovery mechanisms, in the presence of environmental disturbances in both simulation (Sec.~\ref{sec:main-exp-sim}) and real-world environments (Sec.~\ref{sec:main-exp-real})? 2) What are the reasons that lead to the failure cases of AgentChord (Sec.~\ref{sec:more-exp})? 3) Can the data generated by AgentChord, incorporating failure recovery behaviors, be utilized to train a robust policy with failure recovery ability (Sec.~\ref{sec:policy-learning})?

\subsection{Experimental Setup}

\textbf{Hardware setup.} We evaluate AgentChord on a dual-arm robotic platform designed for coordinated bimanual manipulation. The system consists of a fixed-base robot equipped with two 6-DoF arms and parallel-jaw grippers, enabling a range of synchronous and asynchronous dual-arm tasks such as grasping, handovers, and fine manipulation. To support robust perception in dynamic and cluttered environments, we employ a multi-view RGB-D sensing setup with two binocular cameras providing complementary viewpoints. Full hardware specifications are provided in the Appendix~\ref{sec:hardware}.

\textbf{Task Settings.} We evaluate AgentChord on six tasks: 1) \texttt{Single-arm pour water}: Pour water into a cup using a single arm.
2) \texttt{Dual-arm pour water}: Pour water into a cup using both arms.
3) \texttt{Rearrange table}: Rearrange objects such as a fork and spoon beside a plate.
4) \texttt{Handover block}: Handover a block to a specified location.
5) \texttt{Fold towel}: Fold a towel in half.
6) \texttt{Setup coffee tray}: Place a coffee capsule in a tray and move it forward. Further task details are provided in the Appendix~\ref{sec:task-description}.

\begin{figure}[htbp]
    \centering
    \includegraphics[width=\linewidth]{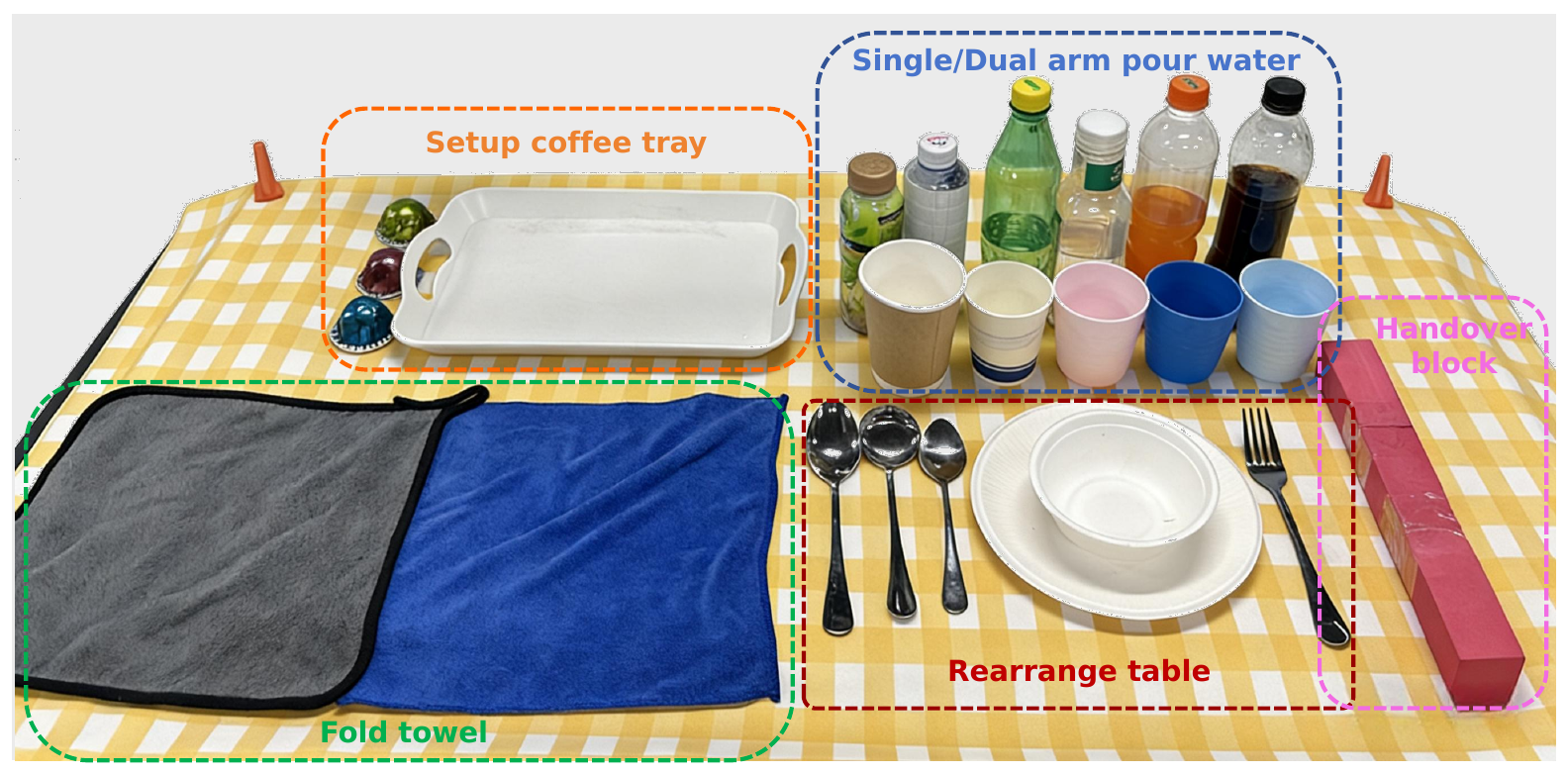}
    \vspace{-0.2in}
    \caption{We collected a variety of object instances for each task to verify the generalizability of different methods.}
    \vspace{-0.1in}
    \label{fig:task-overview}
\end{figure}

Each task consists of 20 trials, in which the manipulated objects are varied across different instances within the same category and their initial poses are randomly sampled. Fig.~\ref{fig:task-overview} shows the related objects in each task.

\textbf{Baseline Methods.} 
We compare AgentChord against four strong baselines that are capable of detecting and recovering from execution failures.
1) Inner Monologue (\textbf{IM})~\cite{huang2022inner} employs VLMs to detect failures at the completion of each sub-goal using a VQA formulation.
2) DoReMi (\textbf{DRM})~\cite{guo2024doremi} detects intermediate failures during execution by issuing frequent VQA-based queries every several steps.
3) \textbf{ReKep} proposes scene keypoints and employs VLM to generate constraint detectors in the form of executable programs, resulting in improved detection efficiency.
4) Code-as-Monitor (\textbf{CaM})~\cite{zhou2025code} trains a constraint painter and tracks the resulting constraints using code scripts for efficient failure monitoring. Notably, IM, DoReMi, and CaM trigger parts of the pipeline after each sub-goal completion or failure detection, while ReKep performs backtracking upon failure detection. For consistency, all methods utilize GPT-5~\cite{singh2025openai}, accessed via API calls, along with a shared set of functional tools.

\textbf{Evaluation Metrics.} We report the \textit{task success rate} as the primary evaluation metric. To evaluate the efficiency of AgentChord, we additionally report the \textit{average execution time} and the \textit{average number of steps}. Execution time is measured from the moment the system receives the initial inputs, including visual observations and task instructions, until the task is completed. This metric captures the full system latency, encompassing perception, planning, action execution, and all MLLM inference costs. The number of steps is defined as the total count of action steps executed to complete a task, including failure recovery actions.

\vspace{-0.05in}
\subsection{Performance in Simulated Manipulation Tasks}~\label{sec:main-exp-sim}
\begin{table*}[htbp]
\centering
\caption{Evaluation results on three simulation tasks with different degrees of disturbances. The bolded values indicate the best results (highest success rate, lowest execution time, and lowest episode steps) for each setting. In cases of tied primary metrics, bolding is determined by the better secondary metrics.}
\label{tab:sim-results}
\small
\vspace{-0.05in}
\resizebox{\textwidth}{!}{%
\begin{tabular}{cc|ccccc|cccc|cccc}
\toprule
\multicolumn{2}{c|}{\multirow{2}{*}{\textbf{Tasks with disturbance}}} 
& \multicolumn{5}{c|}{\textbf{Success Rate (\%) $\uparrow$}} 
& \multicolumn{4}{c|}{\textbf{Execution Time (s) $\downarrow$}} 
& \multicolumn{4}{c}{\textbf{Episode Steps $\downarrow$}} \\
\cmidrule(lr){3-7}\cmidrule(lr){8-11}\cmidrule(lr){12-15}
\multicolumn{2}{c|}{} 
& IM & DRM & ReKep & CaM & AgentChord 
& DRM & ReKep & CaM & AgentChord
& DRM & ReKep & CaM & AgentChord \\
\midrule

\multirow{2}{*}{\begin{tabular}[c]{@{}c@{}}Single-arm pour water\\ with drop $p$\end{tabular}}
& $p$=0.05
& 85 & 95 & 100 & 100 & \textbf{100}
& 96.7 & 38.2 & 72.4 & \textbf{33.1}
& 356 & 362 & \textbf{338} & 340 \\
& $p$=0.10
& 75 & 90 & 95 & 100 & \textbf{100}
& 126.0 & 46.1 & 98.1 & \textbf{38.8}
& 389 & 380 & 361 & \textbf{355} \\

\midrule
\multirow{2}{*}{\begin{tabular}[c]{@{}c@{}}Dual-arm pour water\\ with drop $p$\end{tabular}}
& $p$=0.05
& 75 & 90 & 85 & 95 & \textbf{100}
& 103.8 & 69.9 & 83.9 & \textbf{50.2}
& 418 & 438 & 409 & \textbf{392} \\
& $p$=0.10
& 70 & 80 & 75 & 90 & \textbf{95}
& 145.3 & 93.1 & 110.0 & \textbf{63.4}
& 512 & 531 & 492 & \textbf{483} \\

\midrule
\multirow{2}{*}{\begin{tabular}[c]{@{}c@{}}Rearrange table\\ with drop $p$\end{tabular}}
& $p$=0.05
& 90 & 100 & 100 & 100 & \textbf{100}
& 64.2 & 34.2 & 43.1 & \textbf{29.6}
& 267 & 298 & 257 & \textbf{255} \\
& $p$=0.10
& 80 & 100 & 100 & 100 & \textbf{100}
& 99.2 & 45.0 & 61.2 & \textbf{33.9}
& 296 & 337 & \textbf{280} & 285 \\

\midrule
\multicolumn{2}{c|}{\textbf{Average}}
& 79.2 & 92.5 & 92.5 & 97.5 & \textbf{99.2}
& 105.9 & 54.4 & 78.1 & \textbf{41.5}
& 373 & 391 & 356 & \textbf{352} \\
\bottomrule
\end{tabular}}
\vspace{-0.15in}
\end{table*}
We conduct the experiments in simulation as a preliminary validation, as it provides a more controlled and fair evaluation setting by enabling precise simulation of desired disturbances like object slipping. Moreover, using simulation allows us to eliminate the influence of perception modules by directly accessing ground-truth object states and poses, thereby focusing on the performance of the agentic system itself. In this experiment, we utilize the public robotic simulation platform EmbodiChain~\cite{EmbodiChain}, which integrates GPU-accelerated physics simulation, high-fidelity rendering, modular learning environments, and MLLM-based embodied reasoning agents. We select \texttt{single-arm pour water}, \texttt{dual-arm pour water}, and \texttt{rearrange table} as three representative tasks for evaluation. In these tasks, external disturbances are introduced stochastically according to predefined probabilities: with probability $p$, an object held by the end-effector is randomly dropped at each atomic action.

Quantitative results are summarized in Table~\ref{tab:sim-results}. Overall, AgentChord consistently outperforms baseline methods across all evaluated tasks with disturbances, achieving higher success rates while requiring shorter execution times and fewer episode steps. The superior success rate highlights AgentChord’s strong capability to anticipate potential failures and proactively generate appropriate recovery actions within a well-organized and coordinated agentic framework. Moreover, its efficient failure detection and recovery mechanism enables rapid task completion, resulting in the lowest average execution time and the lowest average episode step count among all methods.

We observe that ReKep~\cite{huang2024rekep} performs poorly on the dual-arm pour water task. This limitation arises because simple backtracking strategies are insufficiently robust for long-horizon bimanual manipulation, which requires both asynchronous and synchronous coordination between arms. Nevertheless, ReKep achieves the second-best execution time overall, as it invokes the full pipeline only once at task initialization and relies on backtracking upon failure detection. While this design avoids additional MLLM invocations and thus reduces inference latency, it often introduces redundant backtracking actions that are not strictly necessary for recovery, which is reflected in its longer episode steps.

In contrast, CaM~\cite{zhou2025code} achieves a similar number of episode steps to AgentChord but incurs noticeably longer execution times. This overhead stems from repeatedly invoking an MLLM upon sub-task completion or failure detection. The performance gap between CaM and AgentChord further indicates that MLLMs can be more effectively leveraged to generate recovery actions in advance during failure anticipation, rather than being re-invoked after failures occur.

Upon failure detection, the system triggers a recovery transition, moves to a recovery node (e.g., re-grasping the fallen cup), and resumes the nominal execution without re-invoking the MLLM. Visualization results are shown in the Appendix~\ref{sec:more-sim-results}.

\subsection{Performance in Real-world Manipulation Tasks}~\label{sec:main-exp-real}
In this section, we conduct real-world experiments on six predefined tasks to provide a more comprehensive and practical evaluation. For each task, we explicitly introduce external disturbances to induce failures, such as perturbing the object during grasping or forcibly removing it from the end-effector during motion. These disturbances are randomly applied once or twice per trial at any point during execution and are kept identical across all compared methods to ensure fairness. It is worth noting that, in the \texttt{rearrange table} and \texttt{fold towel} tasks, failures naturally occur even without external disturbances, as the target objects are very thin or deformable, making them challenging to manipulate.

\begin{table*}[t]
\centering
\caption{Evaluation results on six real-world tasks with disturbances. Bold values indicate the best results. In cases of tied primary metrics, bolding is determined by the better secondary metrics.}
\label{tab:real-results}
\resizebox{0.9\textwidth}{!}{
\begin{tabular}{l|ccccc|cccc}
\toprule
\multirow{2}{*}{\textbf{Tasks with disturbance}} 
& \multicolumn{5}{c|}{\textbf{Success Rate (\%) $\uparrow$}} 
& \multicolumn{4}{c}{\textbf{Execution Time (s) $\downarrow$}} \\
\cmidrule(lr){2-6} \cmidrule(lr){7-10}
& \textbf{IM} & \textbf{DRM} & \textbf{ReKep} & \textbf{CaM} & \textbf{AgentChord} 
& \textbf{DRM} & \textbf{ReKep} & \textbf{CaM} & \textbf{AgentChord} \\
\midrule
Single-arm pour water   & 70 & 80 & 85 & 90 & \textbf{95} & 130.5 & 87.0  & 119.5 & \textbf{75.9} \\
Dual-arm pour water     & 60 & 65 & 60 & 70 & \textbf{80} & 185.0 & 130.8 & 167.2 & \textbf{110.7} \\
Rearrange table         & 80 & 85 & \textbf{95} & 90 & 90 & 113.4 & 85.4 & 99.3 & \textbf{71.0} \\
Handover block          & 60 & 70 & 60 & 75 & \textbf{85} & 178.6 & 149.5 & 170.2 & \textbf{130.1} \\
Fold towel              & 40 & 50 & 50 & 50 & \textbf{55} & 117.5 & 83.7 & 108.1 & \textbf{78.4} \\
Setup coffee tray       & 45 & 50 & 40 & 60 & \textbf{60} & 135.8 & 105.9 & 121.3 & \textbf{87.3} \\
\midrule
\textbf{Average} 
& 59.2 & 66.7 & 65.0 & 72.5 & \textbf{77.5}
& 143.5 & 107.1 & 130.9 & \textbf{92.2} \\
\bottomrule
\end{tabular}
}
\vspace{-0.1in}
\end{table*}

\begin{figure*}[t]
    \centering
\vspace{0.2in}     \includegraphics[width=\linewidth]{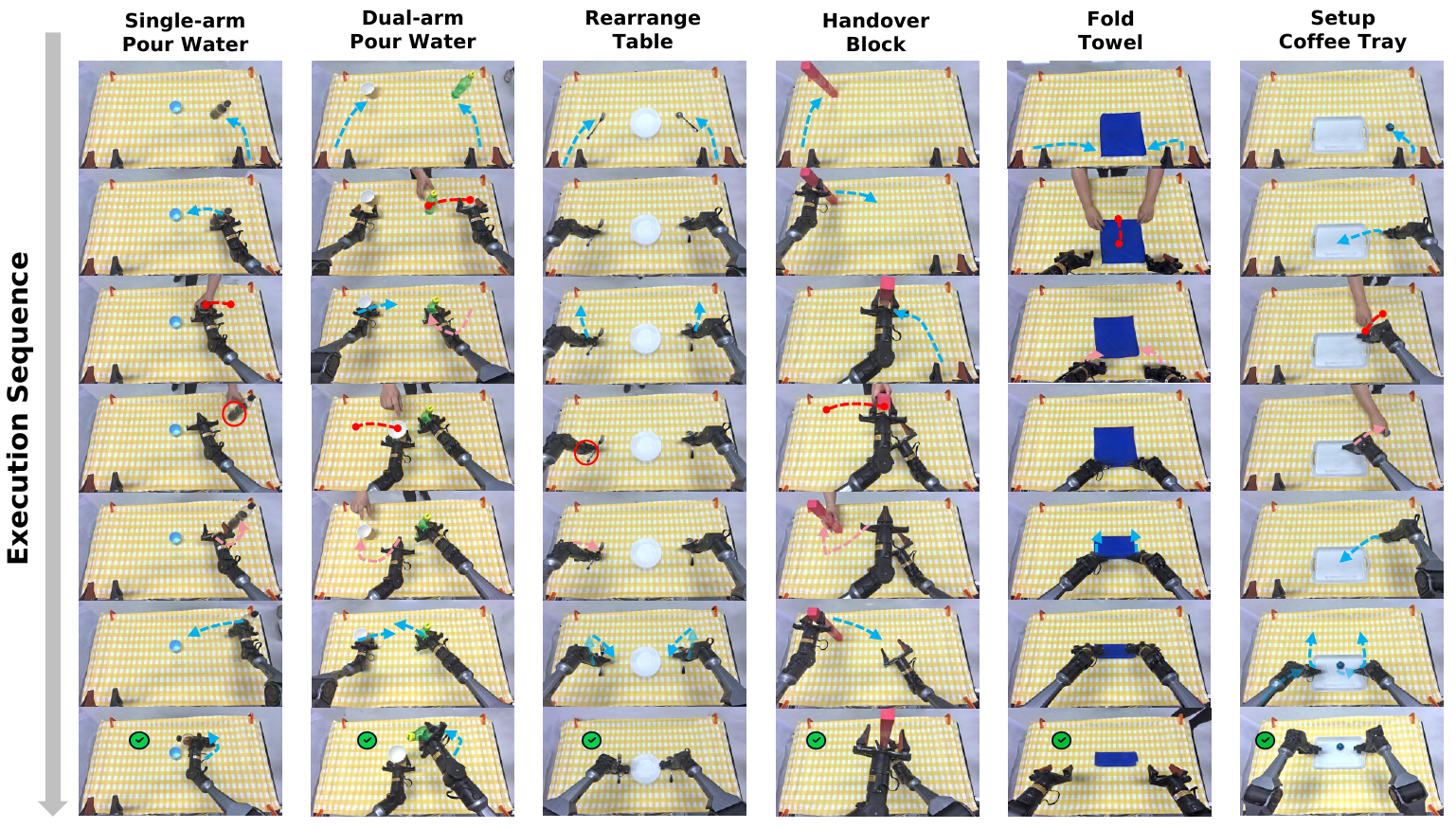}
    \vspace{-0.25in}
    \caption{Illustration of the failure recovery process in AgentChord for six real-world tasks.}
    \label{fig:main-exp}
\end{figure*}

The results are summarized in Table~\ref{tab:real-results}. Because episodic steps are difficult to measure reliably in real-world experiments, we do not report this metric. We can find that AgentChord achieves the best overall performance, attaining the highest average success rate across the six tasks and the shortest execution time. This demonstrates AgentChord’s strong capability for failure anticipation and recovery within a structured graph framework, even under real-world conditions.

Although CaM~\cite{zhou2025code} attains relatively high success rates, it relies heavily on repeated MLLM invocations at each sub-goal and upon each detected failure, leading to substantial computational overhead and latency. ReKep~\cite{huang2024rekep} performs competitively on the \texttt{rearrange table} task, which involves relatively few sub-goals and can be handled effectively through simple backtracking. However, its performance degrades on more complex tasks, as a purely regressive recovery strategy is poorly suited to long-horizon manipulation with both asynchronous and synchronous bimanual coordination.

Fig.~\ref{fig:main-exp} presents the visualization results, which capture a single execution of each task with various failures occurring during different stages of the process. It demonstrates how AgentChord responds to the compiled failure-monitoring functions with the corresponding recovery plan. When a failure is detected, the system immediately follows the augmented recovery edge as shown in pink lines; otherwise, it proceeds along the original graph as shown in blue lines. We provide more visualization results in the Appendix~\ref{sec:more-real-results}.

\begin{figure}[t]
    \centering
    \vspace{0.05in}
    \includegraphics[width=\linewidth]{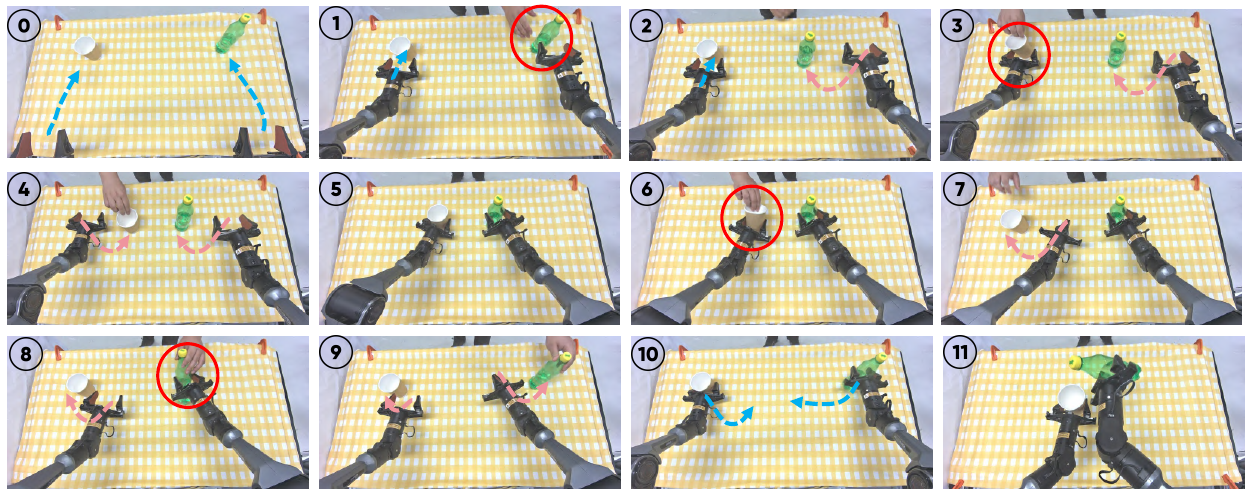}
    \vspace{-0.2in}
    \caption{Failure detection and recovery of AgentChord in dual-arm pour water, where execution switches to a pre-compiled recovery branch and then rejoins the nominal task graph. Visualizations with more tasks and types of disturbances (e.g., tipped object) are presented in the Appendix~\ref{sec:more-real-results}.}
    \label{fig:many-recovery}
    \vspace{-0.05in}
\end{figure}

To evaluate the recovery capability of AgentChord under more severe conditions, we conduct stress tests where frequent external disturbances are introduced. As shown in Fig.~\ref{fig:many-recovery}, the results show the robustness of the proposed method under sustained perturbations.

\subsection{Deep Analysis on the Failure Cases}~\label{sec:more-exp}
Most failures in our real-world experiments arise from limitations in the agent’s reasoning ability, which become more pronounced in long-horizon tasks. In such settings, certain failure modes may not be fully anticipated during graph construction, or the agent may recover to an incorrect task graph node. For example, during a block handover from the left to the right arm, if the block is unintentionally displaced, the agent may fail to reason that the correct recovery strategy is to retreat the right arm, reopen the gripper, and re-initiate the handover sequence. We expect that stronger reasoning models would improve anticipation accuracy and recovery selection.

A second source of failure stems from IK limitations during recovery execution. Even when the agent plans a reasonable corrective behavior, the corresponding IK problem may be infeasible under the current scene. This issue is particularly evident when re-grasping fallen or tipped objects. For instance, if a block is knocked over, the agent may attempt to re-grasp it from an unfavorable orientation, but the IK solver fails to find a valid solution due to joint limits or insufficient reachability. Such failures highlight the gap between high-level recovery intent and low-level kinematic feasibility.

Finally, perception and execution compilation errors also contribute to occasional failures. Inaccurate object masks or noisy point clouds can lead to imprecise pose estimates, which further degrade grasp planning and recovery execution. In addition, the compilation agent may sometimes invoke incorrect function tools for monitoring, resulting in mismatches between the intended recovery logic and the executed checks. For example, to determine whether an object is securely held, monitoring the gripper aperture is often the most reliable strategy. However, in some cases, the agent instead relies on the relative distance between the object and the gripper, which can be unreliable under noisy perception and pose estimation errors. Improving perception robustness and monitoring logic remain important directions for future work.

\subsection{Leveraging Generated Data for Policy Learning}~\label{sec:policy-learning}
Intuitively, the failure-recovery behaviors generated by AgentChord could be valuable for policy learning. In this section, we present a simple yet effective experiment to assess whether incorporating such trajectories can enhance the policy's ability to recover from failures. We focus on the single-arm pour water task in the simulation, specifically targeting the failure scenario where the cup's position is shifted after the right arm has grasped the bottle.

Building on Sim2Real-VLA~\cite{zhao2026simreal}, an affordance-driven vision-language-action model, we adopt its fine-tuned policies as our baselines. For each baseline method, we further fine-tune the policy using 40 successful trajectories generated in the EmbodiChain simulation environment~\cite{EmbodiChain}. This additional fine-tuning is intended only to adapt the policy to minor changes in the environment configuration, such as table height and camera pose, rather than to learn the task from scratch. Therefore, a small number of trajectories is sufficient in this setting. To evaluate whether AgentChord-generated failure-recovery data benefits policy learning, we keep the total amount of fine-tuning data fixed and replace half of the successful trajectories with failure-recoverable trajectories. This variant is denoted as Sim2Real-VLA (rec).

\begin{table}[t]
\vspace{0.05in}
\centering
\caption{Success rates of different policies on the simulated single-arm pour water task. Bold values indicate the best result.}
\label{tab:policy}
\resizebox{0.7\linewidth}{!}{
\begin{tabular}{l|c}
\toprule
\textbf{Method} & \textbf{\makecell{Success Rate}} \\
\midrule
RDT~\cite{liu2024rdt} & 16/50 \\
$\pi_0$~\cite{black2024pi_0} & 20/50 \\
GR00T N1.5~\cite{bjorck2025gr00t} & 15/50 \\
Sim2Real-VLA~\cite{zhao2026simreal} & 26/50 \\
\midrule
\textbf{Sim2Real-VLA (rec)} & \textbf{39/50} \\
\bottomrule
\end{tabular}
}
\vspace{-0.1in}
\end{table}

We evaluate each policy over 50 trials in the simulation environment, where the cup is moved during execution to simulate failure scenarios. The results are presented in Table~\ref{tab:policy}. It indicates that the policy fine-tuned with failure-recoverable trajectories exhibits improved robustness and the ability to recover from the specified failure scenario compared to the baseline policy. This suggests that incorporating recovery behaviors into the training data can significantly improve robotic policy performance, particularly in real-world scenarios where failure recovery is essential for task success. This highlights the importance of AgentChord in automatically generating such trajectories within an agentic system for policy training.

\section{Limitations}
\label{sec:limitations}
AgentChord relies on the recovery orchestration agent to anticipate likely failure modes before execution. Although MLLMs can identify many plausible failures, they may still miss rare, compound, or non-hand-induced failures, especially in long-horizon tasks. If such a failure is not covered by the pre-compiled monitors or recovery branches, the system may not trigger an appropriate recovery behavior. A possible fallback is to invoke an MLLM-based verifier at sub-goal boundaries or upon execution timeout. If the current sub-goal is not completed, the agent pipeline can be re-invoked to diagnose the unforeseen failure and synthesize an additional recovery branch for this previously unpredicted failure mode.

The reliability of failure monitoring depends on the quality of the feature extractor $\Phi(\hat{x}_t,\xi_t)$. Some implemented features, such as PCA-based orientation estimation, assume stable object geometry and may become unreliable for irregular, deformable, or transparent objects with noisy boundaries or ill-defined dominant axes. Nevertheless, this limitation is not inherent to the framework, since $\Phi$ is a modular feature interface. Stronger extractors, such as relational keypoints, MLLM-derived semantic-geometric features, or point-cloud foundation models, can be incorporated without changing the graph construction, monitoring, or recovery formulation.

\section{Conclusion}
In this paper, we presented AgentChord, a choreographed agentic framework for robust robotic manipulation with proactive failure recovery. AgentChord represents each task as a directed graph, where nodes encode semantic sub-goals and edges encode constraint-aware transitions. Before execution, a recovery orchestration agent augments this nominal graph with anticipated failure modes, recovery nodes, and forward-moving recovery branches. An execution compilation agent then instantiates both nominal and recovery transitions using hierarchical constrained solvers and low-latency compiled monitors, allowing the robot to immediately switch to a pre-compiled recovery behavior once a failure is detected.

Experiments on long-horizon single-arm and bimanual manipulation tasks in both simulation and real-world settings show that AgentChord improves task success rates and reduces execution time compared with reactive re-planning or backtracking-based baselines. Additional results further show that the failure-recovery trajectories generated by AgentChord can benefit downstream policy learning. These results suggest that recovery-augmented task graphs provide a practical and extensible representation for integrating failure anticipation, monitoring, and recovery in autonomous robotic manipulation.

\section*{Acknowledgments} This work is supported in part by 
Shenzhen Science and Technology Program under grant KJZD20240903104008012, Shenzhen Science and Technology Program under grant ZDCY20250901113000001, CUHK-CUHK(SZ)-GDSTC Joint Collaboration Fund No. 2025A0505000053, and Guangdong Provincial Key Laboratory of Mathematical Foundations for Artificial Intelligence (2023B1212010001).

\vspace{0.2in}
\bibliographystyle{plainnat}
\bibliography{references}

\appendices
\section{Hardware System}\label{sec:hardware}
\textbf{Dual-Arm Robot.} In this paper, we focus on bimanual robotic manipulation, as it is both more practical and more challenging than single-arm scenarios. We select the AgileX CobotMagic~\cite{fu2024mobile} robot as our manipulation platform, which is equipped with two 6-DoF arms and a fixed base to emphasize coordinated dual-arm actions without the need for mobile base motion. The robot uses two parallel-jaw grippers with a maximum opening distance of 10 cm. This configuration is widely adopted in various research studies~\cite{black2024pi_0, liu2024rdt}. The robot’s dual arms with grippers enable the execution of complex bimanual tasks such as simultaneous grasping, object handovers, and fine manipulation in cluttered environments.

\textbf{Camera Configuration.} Building on the setup in~\cite{huang2024rekep, zhou2025code}, which utilizes two or three RGB-D cameras for more comprehensive scene perception to minimize occlusion in dynamic environments, we also use two RGB-D cameras (a third-person top view and a front view). Each camera is a binocular stereo type, similar to commercial RGB-D cameras, but with the added benefit of providing raw left and right images for greater flexibility in post-processing~\cite{zhou2025yoto}. This configuration improves the robot’s ability to capture detailed spatial information, particularly in complex scenarios. The cameras capture calibrated RGB-D images at a fixed frequency of 10 Hz.

\begin{figure}[ht]
    \centering
    \includegraphics[width=0.8\linewidth]{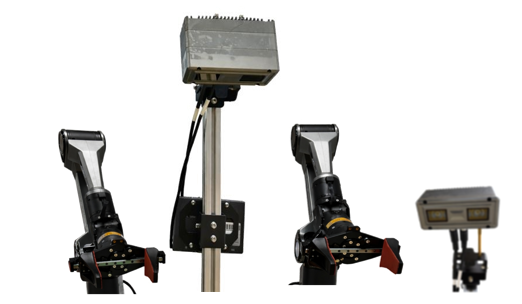}
    \caption{Overview of the hardware system, consisting of an AgileX CobotMagic dual-arm robot and two industrial binocular stereo cameras.}
    \label{fig:device}
\end{figure}

Figure~\ref{fig:device} shows an overview of the used hardware system.

\section{Detailed Task Descriptions}\label{sec:task-description}
We evaluate AgentChord on six tasks: \texttt{single-arm pour water}, \texttt{dual-arm pour water}, \texttt{rearrange table}, \texttt{handover block}, \texttt{fold towel}, and \texttt{setup coffee tray}. L and R denote the left and right arms, respectively.
A task is considered successful only if all task-specific success criteria are satisfied at the end of execution.

\begin{itemize}
\item \texttt{Single-arm pour water}: Pour water from the bottle into a cup using a single arm, including: Grasp bottle (R) $\rightarrow$ Move bottle to cup (R) $\rightarrow$ Pour water (R) $\rightarrow$ Place bottle (R).
\textbf{Success criterion}: The bottle is rotated to a designated pouring orientation above the cup, indicating a successful pouring action, and is subsequently returned to a stable placement on the table.

\item \texttt{Dual-arm pour water}: Pour water from the bottle into the cup using both arms, including: Grasp cup and bottle (LR) $\rightarrow$ Move cup to receiving position (L) $\rightarrow$ Move bottle to cup (R) $\rightarrow$ Pour water (R) $\rightarrow$ Return cup and bottle (LR).
\textbf{Success criterion}: The bottle is rotated to a designated pouring orientation above the cup, and both the cup and bottle are returned to stable, upright placements.

\item \texttt{Rearrange table}: Rearrange the fork and spoon beside the plate, including: Grasp fork and spoon (LR) $\rightarrow$ Lift fork and spoon (LR) $\rightarrow$ Place fork and spoon beside plate (LR).
\textbf{Success criterion}: Both the fork and spoon are placed in their designated target regions beside the plate with correct orientation.

\item \texttt{Handover block}: Handover the block and place it in a specified location, including: Grasp block (L) $\rightarrow$ Move block to handover position (L) $\rightarrow$ Handover block from left to right (LR) $\rightarrow$ Place block (R).
\textbf{Success criterion}: The block is successfully transferred between arms and placed within the target region without being dropped.

\item \texttt{Fold towel}: Fold the towel in half, including: Grasp towel bottom corners (LR) $\rightarrow$ Lift towel bottom (LR) $\rightarrow$ Move to towel top corners (LR) $\rightarrow$ Place towel (LR).
\textbf{Success criterion}: The two bottom corners of the towel are placed within a tolerance of the corresponding top corners, resulting in a half-folded configuration.

\item \texttt{Setup coffee tray}: Place the coffee capsule into the tray and move it forward, including: Grasp coffee capsule (L) $\rightarrow$ Place coffee capsule into tray (L) $\rightarrow$ Grasp tray (LR) $\rightarrow$ Place tray forward (LR).
\textbf{Success criterion}: The coffee capsule is correctly placed inside the tray, and the tray is moved to the designated forward position.
\end{itemize}
These tasks involve both asynchronous and synchronous types of dual-arm collaboration, adding complexity to the system. The objects manipulated can be either rigid or deformable.

\section{More Implementation Details}\label{sec:imp-details}
\subsection{Algorithm}
Algorithm~\ref{alg:agentchord} summarizes the overall AgentChord pipeline. 
Given the task instruction and initial observation, the task agent first constructs a nominal task graph $\mathcal{G}=(V,E)$, where nodes represent semantic sub-goals and edges represent constraint-aware transitions. 
The recovery agent then analyzes each nominal edge to anticipate likely failure modes, generate the corresponding monitoring functions, and augment the graph with recovery branches and recovery mappings. 
After filtering recovery branches according to the forward-moving principle, the compilation agent instantiates node keyframes, edge programs, and monitors using the constrained solvers. 
During online execution, AgentChord follows the active edge program while continuously updating the feature vector $z_t$. 
Once a monitor is triggered, execution immediately switches to the mapped recovery edge, enabling low-latency recovery without re-planning.

\begin{algorithm}[t]
\caption{AgentChord Pipeline}
\label{alg:agentchord}
\begin{algorithmic}[1]
\Require instruction $\mathcal{I}$, initial observation $o_0$, tools 
$\mathcal{F}_{\text{perc}}, \mathcal{F}_{\text{action}}, \mathcal{F}_{\text{mllm}}$
\Statex // \textit{Task Agent: build nominal graph $\mathcal{G}$}
\State $\mathcal{G}=(V,E)\leftarrow \textsc{Structure}(\mathcal{I},o_0;\mathcal{F}_{\text{mllm}})$
\Statex // \textit{Recovery Agent: augment with recovery branches}
\For{each nominal edge $\varepsilon^{i\rightarrow j}\in E$}
    \State $\mathcal{F}^{i\rightarrow j}\leftarrow
    \textsc{PredictFailures}(\mathcal{I},o_0,\mathcal{G},\varepsilon^{i\rightarrow j};\mathcal{F}_{\text{mllm}})$
    \State Generate recovery nodes, recovery edges, monitoring functions $f_m^{i\rightarrow j}(z_t)$, and mappings
    $\rho(\varepsilon^{i\rightarrow j},f_m^{i\rightarrow j})$
\EndFor

\State Construct $\mathcal{G}_{\text{aug}}$ and filter recovery branches by Eq.~(\ref{eq:forward_progress})
\Statex // \textit{Compilation Agent: compile graph to executable programs}
\State Compile node keyframes $\{e^i\}$ using Eq.~(\ref{eq:subgoal_solver})
\State Compile edge programs $\{\pi^{i\rightarrow j}\}$ using Eq.~(\ref{eq:path_solver})
\State Compile monitors $\{\mathcal{M}^{i\rightarrow j}_m\}$ using Eq.~(\ref{eq:monitor})
\Statex // \textit{Online execution: run edge programs; monitors trigger low-latency switching}
\State $v\leftarrow v^0$
\While{$v\neq v^N$}
    \State Select active edge $\varepsilon^{i\rightarrow j}\in\textsc{Out}(v)$
    \While{not $\textsc{Satisfy}(\mathcal{C}_{\text{sub}}^j)$}
        \State $o_t \gets \textsc{Sense}()$, $\hat{x}_t \gets \Psi(o_t)$,
        $\xi_t\gets(q_t,g_t)$,
        $z_t\gets\Phi(\hat{x}_t,\xi_t)$ using $\mathcal{F}_{\text{perc}}$
        \State Execute one action chunk of $\pi^{i\rightarrow j}$ using $\mathcal{F}_{\text{action}}$
        \If{$\exists m$ such that $\mathcal{M}^{i\rightarrow j}_m(t)=1$}
            \State $\varepsilon^{i\rightarrow j}\leftarrow
            \rho(\varepsilon^{i\rightarrow j},f_m^{i\rightarrow j})$
        \EndIf
    \EndWhile
    \State $v\leftarrow v^j$
\EndWhile
\end{algorithmic}
\end{algorithm}

\subsection{Robot Control}
AgentChord executes manipulation behaviors through a library of predefined \emph{atomic actions} (as summarized in Table~\ref{tab:atom-actions}). Each atomic action represents a semantically meaningful skill and is internally realized as a sequence of low-level robot joint commands as follows.

\textbf{End-effector pose specification.} Each atomic action first specifies one or more target 6-DoF end-effector poses expressed in the world frame, optionally invoking perception modules to estimate object poses or reference frames when required. These target poses may be defined absolutely (e.g., move to a fixed pose) or relatively (e.g., move by an offset with respect to an object frame), depending on the action semantics. Before execution, all target poses are clipped to a predefined workspace to ensure kinematic feasibility and safety.

\textbf{Inverse kinematics and interpolation.} Given a target end-effector pose, we compute the corresponding joint configuration using the inverse kinematics solvers provided by {EmbodiChain}~\cite{EmbodiChain}, which can incorporate the constraint solvers as described in Appendix~\ref{sec:appendix_subgoal_solver} and Appendix \ref{sec:appendix_path_solver}. To ensure smooth execution, EmbodiChain performs joint-space interpolation between the current and target joint configurations with predefined step sizes. Each interpolated joint configuration is then issued as a control target, producing a sequence of joint commands that collectively form the execution trajectory of the atomic action.

\textbf{Low-level control.} The interpolated joint commands are executed using position control by the robot. This design allows atomic actions to be executed in a stable, smooth, and hardware-compatible manner, while remaining agnostic to the specific robot embodiment. Since all actions are ultimately represented as joint trajectories, both single-arm and bimanual behaviors can be handled in a unified control framework.

\subsection{Atomic Actions}\label{sec:atom-actions}
Table~\ref{tab:atom-actions} summarizes the atomic actions used in this work. Each atomic action can optionally invoke perception modules $\mathcal{F}_{\text{perc}}$ to infer task-relevant targets and plan motions conditioned on the current scene. For example, when executing \texttt{grasp}, the system automatically detects whether the target object is upright or fallen; if the object has fallen, the action adapts by re-grasping and lifting it to a stable configuration. In addition, the atomic action library is extensible and can be augmented with new skills as needed for more complex tasks (e.g., twisting or articulated object manipulation).

\begin{table}[htbp]
\centering
\footnotesize
\setlength{\tabcolsep}{4pt}
\renewcommand{\arraystretch}{1.12}
\caption{Atomic actions used to compose task-graph edges. Each action is parameterized by \texttt{robot\_name} and (when applicable) \texttt{obj\_name}.}\label{tab:atom-actions}
\begin{tabular}{p{0.28\linewidth} p{0.66\linewidth}}
\hline
\textbf{Atomic action} & \textbf{Brief description} \\
\hline
\texttt{drive} &
Dual-arm wrapper executing \texttt{left\_arm\_action} and \texttt{right\_arm\_action} (either can be \texttt{None}). \\

\texttt{grasp} &
Approach and grasp a target object with a pre-grasp offset (e.g., \texttt{pre\_grasp\_dis}); supports optional grasp offsets for alignment. \\

\texttt{open\_gripper} &
Open the gripper to release; optional \texttt{sample\_num} controls actuation steps. \\

\texttt{close\_gripper} &
Close the gripper to secure a hold; optional \texttt{sample\_num} controls actuation steps. \\

\texttt{move\_to\_obj} &
Move to a pose defined by offsets relative to a referenced object (e.g., \texttt{x\_offset}, \texttt{y\_offset}, \texttt{z\_offset}); supports orientation hints and mask-conditioned targeting. \\

\texttt{move\_to\_target} &
Move to an absolute target pose (extrinsic frame). \\

\texttt{move\_by\_offset} &
Apply a relative Cartesian displacement in intrinsic/extrinsic frame (e.g., \texttt{dx}, \texttt{dy}, \texttt{dz}) or angles. \\

\texttt{rotate\_eef} &
Rotate the end-effector by a specified angle; used for tilt-to-pour and restoring upright orientation. \\

\texttt{place} &
Place a grasped object at a target table location \texttt{(x,y)} with optional \texttt{z\_offset}, then release. \\

\texttt{back} &
Return the arm to a predefined home configuration. \\
\hline
\end{tabular}
\end{table}

\subsection{Implementation Details of Geometric Feature Extraction}\label{sec:appendix_geo_feat}
Given the perception output $\hat{x}_t = \Psi(o_t)$ and the robot state $\xi_t$, we compute the geometric feature vector $z_t = \Phi(\hat{x}_t, \xi_t)$ as defined in Eq.~(\ref{eq:rel_feature}). The feature extraction pipeline consists of the following steps.

\textbf{Object Segmentation and Tracking.}
We apply open-vocabulary segmentation using SAM3~\cite{carion2025sam} on the RGB-D observation $o_t$ to obtain a binary mask $m^k_t$ for each target object. By querying the same object label or description across frames, SAM3 provides consistent object tracking (optionally using the mask at time $t$ as a prior for $t{+}1$). The 2D mask is fused with depth data to reconstruct a 3D point cloud $\mathcal{P}^k_t$ in the robot frame.

\textbf{Geometric Attribute Computation.}
From $\mathcal{P}^k_t$, we extract task-relevant geometric features, including the object centroid and characteristic boundary points (e.g., extremal or top/bottom points). The object’s principal orientation is estimated via Principal Component Analysis (PCA), using the dominant axis of the point cloud. When needed, task-specific fractional points (e.g., a fixed-height position along a bottle) are also computed. These attributes are selected to align with the active constraints, such as orientation checks for uprightness or vertical positions for relative height constraints.

\textbf{Integration of Robot State.}
We append the robot’s proprioceptive state $\xi_t = (q_t, g_t)$ to $z_t$, where $g_t$ denotes the gripper opening width or open/closed state. This captures the interaction status (e.g., whether an object is grasped) and supports constraints involving the gripper.

These features constitute a template library of hand-designed functional tools for perception-based feature extraction, which are provided to the LLM to support constraint generation. The output of this pipeline is a feature vector $z_t$ that encodes the geometric state of the scene and the robot. For each relevant object, $z_t$ includes its position, size, and orientation, as well as any relational quantities required by the active constraints. For example, alignment constraints use object orientation vectors to compute angular deviations, while distance constraints rely on object centroids or boundary points and target locations. These features are used by the monitoring module to efficiently evaluate constraint functions.

\subsection{Implementation Details of Sub-Goal Solver}\label{sec:appendix_subgoal_solver} 
For each semantic sub-goal node $v^i$, we compute a target end-effector pose $e^i$ that satisfies the sub-goal constraints $\mathcal{C}^i_{\text{sub}}$ by solving the constrained optimization problem in Eq.~(\ref{eq:subgoal_solver}). Specifically, we seek an $e^i$ that minimizes a task-specific cost $\lambda^i_{\text{sub}}(e; \hat{x}^i)$. Following~\cite{huang2024rekep}, we implement this solver using a hybrid global-local optimization strategy.

\textbf{Initialization and Solver Strategy.} 
We initialize the end-effector pose using the robot’s current (or last attained) pose. Due to the non-convexity of inverse kinematics and geometric constraints, we first perform a coarse global search (e.g., Dual Annealing~\cite{xiang1997generalized}) to explore feasible regions. The best candidate is then refined using a local optimizer such as SLSQP~\cite{kraft1988software} to obtain a high-precision solution.

\textbf{Objective Function Design.} The objective $\lambda^i_{\text{sub}}(e; \hat{x}^i)$ is designed to include soft penalties that guide the optimizer toward preferred solutions, beyond just satisfying the hard constraints. In our implementation, $\lambda^i_{\text{sub}}$ includes: 

\begin{itemize}
    \item \textbf{Collision avoidance.} We penalize configurations that bring the gripper or (if grasped) object points
    close to the scene surface. Practically, we query a voxelized distance field (ESDF/SDF) built from fused depth
    and compute hinge penalties when distances fall below a safety margin.
    \item \textbf{Reachability and IK residual.} Since the decision variable is task-space pose(s), we include a
    reachability term that penalizes IK failures. Given $e$, we compute $q(e)$ using an IK solver initialized from
    the current joint state and penalize residuals or invalid solutions.
    \item \textbf{Pose regularization.} We add a small term $\|e-e_{\text{ref}}\|^2$, encouraging solutions near the
    current pose (or a task-provided reference) to reduce unnecessary motion.
    \item \textbf{Consistency.} To suppress jitter induced by perception noise, we add $\|e-e^{i}_{\text{prev}}\|^2$
    where $e^{i}_{\text{prev}}$ is the most recent solution for node $v^i$.
    \item \textbf{Bimanual Coordination.} We add regularizers on relative arm
    arrangement (e.g., inter-wrist distance bounds) and penalize close distances between left/right arm point sets,
    reducing self-collision risk during constrained keyframes.
\end{itemize}

\subsection{Implementation Details of Path Solver}\label{sec:appendix_path_solver}

The path solver computes a feasible motion plan that transitions the robot between consecutive sub-goals under the specified path constraints. We formulate this as a constrained optimal control problem, which is solved in a receding-horizon fashion as described by Eq. (\ref{eq:path_solver}). 

\textbf{Receding-Horizon Optimization.} Rather than computing an entire open-loop trajectory at once, we use a receding horizon strategy for online path generation. At each control timestep $t$, we solve an optimization over a horizon of $H$ future steps, producing a sequence $\{e_t, e_{t+1}, \dots, e_{t+H}\}$ that moves the end-effector from its current state toward the goal $e^j$ while respecting all constraints at every intermediate step. We then execute the first $M$ steps of this plan on the robot, where $1 \!\le\! M \!\le \!H$ is a shorter execution window (for example, $M\!=\!1$ corresponds to updating the plan at every single time-step). After those $M$ steps, we obtain a new observation $\hat{x}_{t+M}$ and repeat the optimization for the next horizon. In our implementation, $H$ is chosen such that the horizon covers the remainder of the current edge (or a few seconds of motion), and $M$ is set based on the control frequency (we set $M\!=\!5$ in this paper). This receding horizon approach means the path is continuously being refined: at any given moment, the solver has an up-to-date plan that accounts for the latest robot state and perception, which makes the execution robust to disturbances. 

\textbf{Path Objective Terms.} The cost function $\lambda^{i\to j}_{\text{path}}$ is designed to produce smooth, efficient, and safe motions. In our implementation, we include the following key terms: 
\begin{itemize}
    \item \textbf{Smoothness.} Penalize pose increments $\sum_{h=1}^{H}\|e_{t+h}-e_{t+h-1}\|^2$ to avoid jerky motion.
    \item \textbf{Safety margins.} Add collision-distance hinge penalties along the horizon using the same geometric
    engine as the sub-goal solver (e.g., SDF/ESDF queries on gripper/object point samples).
    \item \textbf{Reachability.} Penalize infeasible poses using IK residuals for $q(e_{t+h})$ to avoid sending
    unreachable commands.
\end{itemize}

\subsection{Implementation Details of Proactive Failure Anticipation and Recovery}\label{sec:appendix_failure_recovery}

In AgentChord, path constraints and failure-monitoring functions play complementary roles. 
For an active transition edge $\varepsilon^{i\rightarrow j}$, the path constraints 
$\mathcal{C}^{i\rightarrow j}_{\text{path}}$ specify the conditions that should remain valid during edge execution and are used by the receding-horizon path solver. 
In addition, the recovery orchestration agent predicts edge-specific failure modes 
$\mathcal{F}^{i\rightarrow j}=\{f^{i\rightarrow j}_1,f^{i\rightarrow j}_2,\dots\}$, where each 
$f^{i\rightarrow j}_m$ is implemented as a scalar violation function over the online feature vector 
$z_t=\Phi(\hat{x}_t,\xi_t)$. By convention, $f^{i\rightarrow j}_m(z_t)\le 0$ indicates normal execution with respect to this failure mode, while $f^{i\rightarrow j}_m(z_t)>0$ indicates that the corresponding failure condition has been detected. 
These failure-monitoring functions can be derived from, or semantically aligned with, the path constraints of the active edge, but are maintained separately to support failure-specific recovery routing.

\textbf{Failure monitoring.}
While executing $\varepsilon^{i\rightarrow j}$, AgentChord continuously evaluates the failure-monitoring functions associated with the active edge. 
A recovery is triggered only when a violation remains persistent over a short window according to Eq.~(\ref{eq:monitor}), which reduces spurious activations caused by transient perception noise or short-lived contact fluctuations. 
In practice, $f^{i\rightarrow j}_m$ can be instantiated using task-relevant checks such as the following. 
For readability, we omit the superscript $i\rightarrow j$ in the examples below.

\begin{itemize}
    \item \textbf{Object pose drift / shifted object.}
    For an object $k$ that should remain near a reference position $\bar{p}^k$ during an edge:
    \begin{equation}
    f_{\text{shift}}^k(z_t)=\|p^k_t-\bar{p}^k\|_2-\delta_{\text{shift}}\le 0,
    \end{equation}
    where $\bar{p}^k$ can be taken from the last stable estimate at node $v^i$ or from the expected in-hand pose, and $\delta_{\text{shift}}$ is a tolerance.

    \item \textbf{Tilted object / uprightness violation.}
    For an object expected to remain upright, using its principal axis $u^k_t$ and the world gravity direction $\hat{\mathbf{g}}$:
    \begin{equation}
    f_{\text{tilt}}^k(z_t)=
    \arccos\!\left(\left|{u^k_t}^{\top}\hat{\mathbf{g}}\right|\right)-\theta_{\max}\le 0.
    \end{equation}

    \item \textbf{Not fully grasped / slipped grasp.}
    Using the gripper opening $g_t$ and object-to-gripper proximity:
    \begin{align}
    f_{\text{grasp}}(z_t) &= g_t - g_{\text{th}} \le 0, \\
    f_{\text{attach}}^k(z_t) &= \|p^k_t - p^{\text{grip}}_t\|_2 - \delta_{\text{attach}} \le 0,
    \end{align}
    where $p^{\text{grip}}_t$ denotes the gripper frame origin or fingertip midpoint.

    \item \textbf{Visibility / tracking loss.}
    If the segmentation quality drops below a minimum point count:
    \begin{equation}
    f_{\text{vis}}^k(z_t)=N_{\min}-N^k_t\le 0,
    \end{equation}
    where $N^k_t$ is the number of valid points in $\mathcal{P}^k_t$ after depth fusion.

    \item \textbf{Relational misalignment.}
    For edges that require maintaining a spatial relationship between two entities, such as bottle-cup alignment:
    \begin{equation}
    f_{\text{rel}}^{k,l}(z_t)
    =
    d_{\text{rel}}^{k,l}-\delta_{\text{rel}}\le 0,
    \end{equation}
    where $d_{\text{rel}}^{k,l}$ measures task-specific relative geometry, such as lateral offset or projected alignment error.
\end{itemize}

It is worth noting that the monitor is not highly sensitive to these tolerance parameters. Across tasks, we use shared default settings rather than task-specific tuning. These defaults provide a practical balance between robustness and responsiveness: larger tolerances reduce false alarms caused by perception or execution noise, whereas smaller tolerances enable earlier failure detection. In addition, these thresholds can be automatically adjusted according to the task geometry and scene context during MLLM-based compilation. The aforementioned monitor functions are provided as example templates in the prompt, allowing the MLLM to either reuse them directly or synthesize new failure detectors by composing the available geometric feature extractors described in Appendix~\ref{sec:appendix_geo_feat}.

\section{More Results in Simulation Experiments}\label{sec:more-sim-results}
\begin{figure*}[ht]
    \centering
    \includegraphics[width=\linewidth]{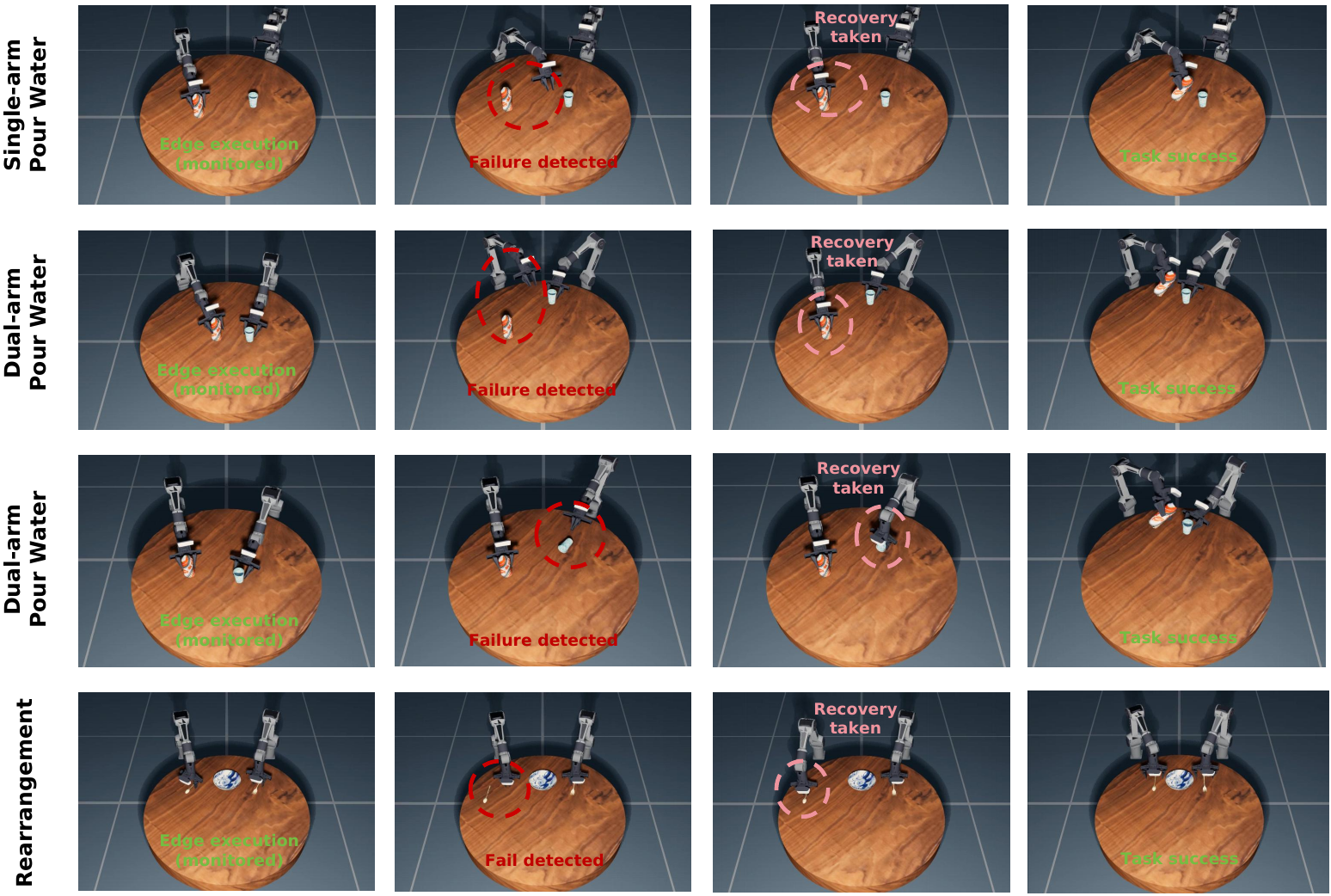}
    \caption{Illustration of failure detection and recovery of AgentChord in three simulation tasks.}
    \label{fig:sim-demo}
    \vspace{-0.15in}
\end{figure*}
Figure~\ref{fig:sim-demo} presents representative illustrations from three simulated tasks. The results demonstrate that AgentChord can proactively anticipate potential failures (e.g., object displacement or slippage) at the initial planning stage, even when such failures may occur at different phases of execution. Upon detection, the system seamlessly transitions through recovery branches and rejoins the nominal task graph, enabling successful task completion.

\section{More Results in Real-world Experiments}\label{sec:more-real-results}
In this section, we present additional visualizations from the real-world experiments. Qualitative videos are provided on the project page.

\subsection{Manipulate Different Objects with Randomness}
\begin{figure*}[ht]
    \centering
    \includegraphics[width=\linewidth]{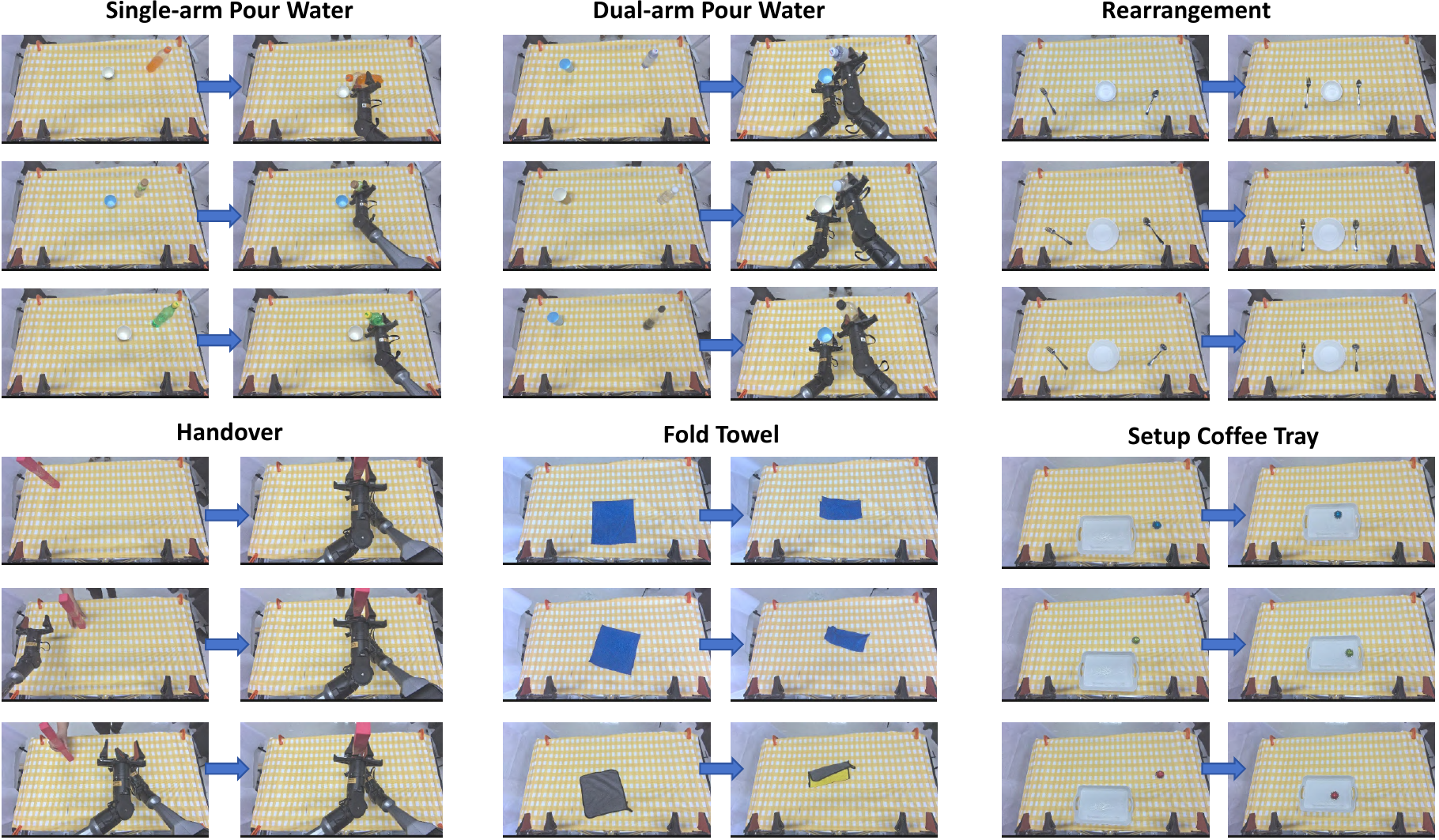}
    \caption{Illustration of the initial and task-completion frames of AgentChord across six real-world tasks. Across different trials, the object instances, poses, and disturbance configurations vary, yet AgentChord consistently completes the tasks successfully.}
    \label{fig:real-demo-random}
    \vspace{-0.15in}
\end{figure*}

Figure~\ref{fig:real-demo-random} visualizes the initial and task-completion frames of AgentChord across six real-world manipulation tasks. In each task, each row corresponds to a different trial, while the two columns show the initial scene configuration and the final successful outcome, respectively. Across trials, the object instances, poses (e.g., position and orientation), and external disturbance configurations vary, within ranges that preserve the kinematic feasibility of the robot. Nevertheless, AgentChord consistently adapts its execution and completes the tasks successfully, demonstrating robustness to environmental variation and real-world uncertainties.

\subsection{Manipulate in Clutter Scenes}
\begin{figure*}[ht]
    \centering
    \includegraphics[width=\linewidth]{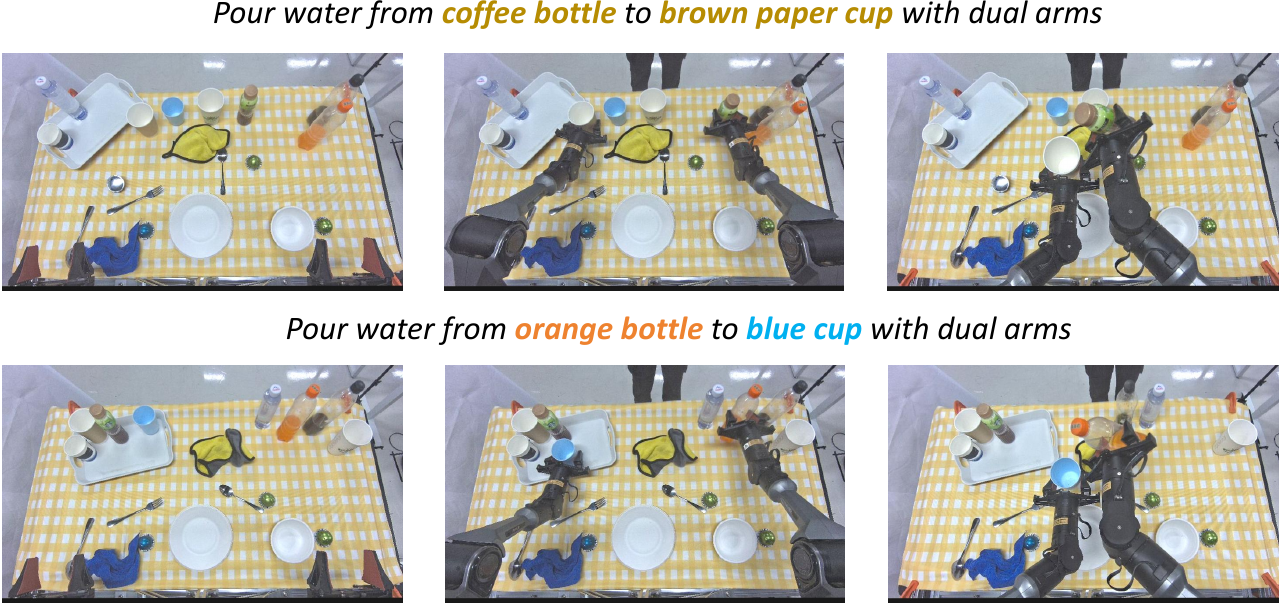}
    \caption{Illustration of AgentChord on the dual-arm pour water task in a cluttered environment.}
    \label{fig:real-demo-clutter}
    \vspace{-0.15in}
\end{figure*}

Figure~\ref{fig:real-demo-clutter} demonstrates AgentChord executing the dual-arm pour water task under cluttered tabletop conditions with varying object instances and target containers. Despite differences in bottle types, cup materials, colors, and spatial arrangements across trials, AgentChord successfully coordinates both arms to stabilize the receiving cup while aligning and tilting the bottle for pouring. These results highlight the system’s robustness to object diversity and environmental clutter, as well as its ability to maintain consistent task execution through coordinated bimanual control and adaptive planning. Each trial is coupled with multiple different disturbances, although we omit the disturbance frames due to space constraints.

\subsection{Ablation on Forward-Moving Recovery Graph}
\label{sec:appendix_ablation_recovery_graph}

To isolate the contribution of the recovery-augmented graph, we compare AgentChord with a backtracking variant, denoted as AgentChord-BT. AgentChord-BT uses the same task graph, feature extractors, and compiled monitors as AgentChord, but removes the pre-compiled recovery nodes and recovery edges. When a monitor is triggered, AgentChord-BT simply backtracks to the previous nominal node and re-executes the corresponding nominal edge, rather than following a dedicated forward-moving recovery branch. This variant directly tests whether proactively generating $V_{\text{rec}}$ and $E_{\text{rec}}$ is beneficial beyond compiled monitoring alone.

\begin{table}[htbp]
\centering
\caption{Ablation study on the recovery-augmented graph. We use AC to denote AgentChord and AC-BT to denote its backtracking variant without pre-compiled recovery edges.}
\label{tab:ablation-recovery-graph}
\resizebox{\linewidth}{!}{
\begin{tabular}{l ccc ccc}
\toprule
\multirow{2}{*}{\textbf{Task}} 
& \multicolumn{3}{c}{\textbf{Success Rate (\%) $\uparrow$}} 
& \multicolumn{3}{c}{\textbf{Execution Time (s) $\downarrow$}} \\
\cmidrule(lr){2-4} \cmidrule(lr){5-7}
& \textbf{ReKep} & \textbf{AC-BT} & \textbf{AC} 
& \textbf{ReKep} & \textbf{AC-BT} & \textbf{AC} \\
\midrule
Single-arm pour water & 85 & 85 & \bfseries 95 & 87.0  & 92.3  & \bfseries 75.9 \\
Dual-arm pour water   & 60 & 65 & \bfseries 80 & 130.8 & 139.5 & \bfseries 110.7 \\
Handover block        & 60 & 70 & \bfseries 85 & 149.5 & 153.4 & \bfseries 130.1 \\
\midrule
Average               & 68.3 & 73.3 & \bfseries 86.7 & 122.4 & 128.4 & \bfseries 105.6 \\
\bottomrule
\end{tabular}
}
\end{table}

As shown in Table~\ref{tab:ablation-recovery-graph}, AgentChord-BT slightly outperforms ReKep in success rate, likely due to AgentChord's more flexible feature extraction and compiled monitoring interface. However, AgentChord-BT incurs higher execution time because failures are handled by regressive backtracking and repeated nominal execution. In contrast, AgentChord achieves both higher success rates and lower execution time. Although constructing the recovery-augmented graph introduces additional upfront reasoning cost, the pre-compiled forward-moving recovery branches reduce redundant execution after failures and lead to better overall efficiency.

\section{Prompts for Agents and Example Outputs}\label{sec:prompts}
The prompts used in AgentChord and representative example outputs are provided in the code repository: \url{https://github.com/Jasonxu1225/AgentChord}.
\end{document}